\documentclass[11pt]{article}

\usepackage[final]{acl}

\usepackage{times}
\usepackage{latexsym}

\usepackage[T1]{fontenc}

\usepackage[utf8]{inputenc}
\usepackage{microtype}

\usepackage{inconsolata}

\usepackage{graphicx,url}

\usepackage{xcolor}     
\usepackage{soul}

\usepackage[most]{tcolorbox}
\tcbuselibrary{breakable, skins}  %
\usepackage{booktabs}
\usepackage[normalem]{ulem}
\usepackage{amssymb}
\usepackage{multirow}
\usepackage{enumitem}

\tcbuselibrary{listings,breakable}
\newtcblisting{promptbox}[1][]{
  listing only,
  breakable,
  colback=gray!4,
  colframe=gray!35,
coltitle=black,
colbacktitle=gray!15,
  boxrule=0.3pt,
  arc=1pt,
  left=2pt,
  right=2pt,
  top=2pt,
  bottom=2pt,
  fonttitle=\bfseries\footnotesize,
  title={#1},
  listing options={
    basicstyle=\ttfamily\scriptsize,
    breaklines=true,
    breakatwhitespace=false,
    columns=fullflexible,
    keepspaces=true,
    showstringspaces=false,
    tabsize=2,
    literate={—}{{---}}1 {→}{{$\rightarrow$}}1 {…}{{\ldots}}1 {≈}{{$\approx$}}1,
  }
}
\title{Hallucinations on the Board: \\Tool-Augmented Evaluation of LLM Chess Commentary}

\author{S. Ashwin Hebbar \\
  Princeton University\\
  \texttt{hebbar@princeton.edu}\And
  Peiyao Sheng \\
  Sentient Labs\\
  \texttt{peiyao@sentient.xyz}\AND
Sewoong Oh \\
  University of Washington\\
  \texttt{sewoong@cs.washington.edu}\And
  Pramod Viswanath \\
  Princeton University\\
  \texttt{pramodv@princeton.edu}
}

\usepackage{xcolor}
\usepackage[normalem]{ulem}
\usepackage{listings, amsmath}
 
\newif\ifrevision
\revisionfalse
 
\ifrevision
  \colorlet{revcolor}{blue}
\else
  \colorlet{revcolor}{black}
\fi

\ifrevision
  \newcommand{\del}[1]{\textcolor{red!70!black}{\sout{#1}}}
  \newcommand{\note}[1]{\textcolor{orange}{\textbf{[NOTE: #1]}}}
\else
  \makeatletter
  \newcommand{\del}[1]{\@bsphack\@esphack}
  \newcommand{\note}[1]{\@bsphack\@esphack}
  \makeatother
\fi
 
\newenvironment{revblock}{\par\color{revcolor}\ignorespaces}{\par}

\newcommand{\revcol}{\color{revcolor}}
 
\lstdefinestyle{rev}{
  basicstyle=\scriptsize\ttfamily\color{revcolor},
  breaklines=true,
  frame=single,
  rulecolor=\color{revcolor},
  columns=fullflexible,
  keepspaces=true,
  literate={—}{{---}}1 {→}{{$\rightarrow$}}1 {…}{{\ldots}}1 {≈}{{$\approx$}}1
}

\newif\ifcamready
\camreadyfalse
\ifcamready
  \colorlet{crcolor}{blue}
  \newcommand{\camdel}[1]{\textcolor{red!70!black}{\sout{#1}}}
\else
  \colorlet{crcolor}{black}
  \makeatletter\newcommand{\camdel}[1]{\@bsphack\@esphack}\makeatother%
\fi

\newcommand{\fs}{/\hspace{0pt}\allowbreak}
\newcommand{\fen}[1]{{\ttfamily\scriptsize #1}}

\hypersetup{
  pdftitle={Hallucinations on the Board: Tool-Augmented Evaluation of LLM Chess Commentary},
  pdfauthor={S. Ashwin Hebbar, Peiyao Sheng, Sewoong Oh, Pramod Viswanath}
}

\begin{document}
\maketitle
\begin{abstract}

Superhuman game engines in domains like chess have made expert-level evaluations easily accessible, yet they communicate what is true without the natural-language explanations that make such expertise educationally useful to experts and non-experts alike. Large language models could, in principle, bridge this gap, but they frequently hallucinate due to limited domain-specific knowledge, and standard reference-based or LLM-as-a-judge frameworks cannot reliably detect these errors. In this work, we present ACT-Eval, an evaluation framework that decomposes chess commentary into atomic claims and routes them to engine-supported tools and expert-annotated gold references to assess factual correctness, conceptual coverage, and move-quality judgment. We release a benchmark of 325 position–move pairs spanning pedagogical, tournament, and critical positions, including 125 positions with expert-verified gold atoms and a five-class error taxonomy. Evaluating leading proprietary and open-weight models, we find that factual hallucinations remain pervasive in chess commentary: GPT-5.4 without tools produces incorrect sub-claims 22.0\% of the time, while smaller open-weight models exceed 40\%. Although tool augmentation substantially improves factual correctness and move-quality assessment, coverage of expert strategic and tactical ideas remains limited across all models. Human validation shows that ACT-Eval’s factual judgments fall within the observed range of inter-human agreement, while its coverage scores correlate strongly with human assessments of strategic completeness.
\end{abstract}

\section{Introduction}

Chess has long served as a canonical domain for studying strategic reasoning. Its appeal lies not only in the combinatorial complexity of the game, but also in the rich explanatory tradition that has developed around it. For centuries, chess understanding has been communicated through annotated games, books, lectures, broadcasts, and, more recently, online platforms such as YouTube and Twitch. Strong players make moves intelligible by translating them into plans, threats, tactical motifs, and long-term tradeoffs. The enduring popularity of these formats reflects a simple reality: players and spectators do not only want to know which move is best, but why it works.

Superhuman engines such as Stockfish~\cite{stockfish} and AlphaZero~\cite{silver2018general} have made expert-level evaluation broadly accessible: even a novice can consult an engine to instantly know whether a move is a blunder or a brilliancy. However, they only provide an objective score, which for beginner and intermediate-level players may not mean much more than determining who is better; even elite grandmaster preparation involves spending hours interpreting engine play into human explanations and practical plans. Engines have largely solved the problem of finding strong moves, but not the problem of explaining them.
\looseness=-1

\begin{figure*}
    \centering
    \includegraphics[width=\linewidth]{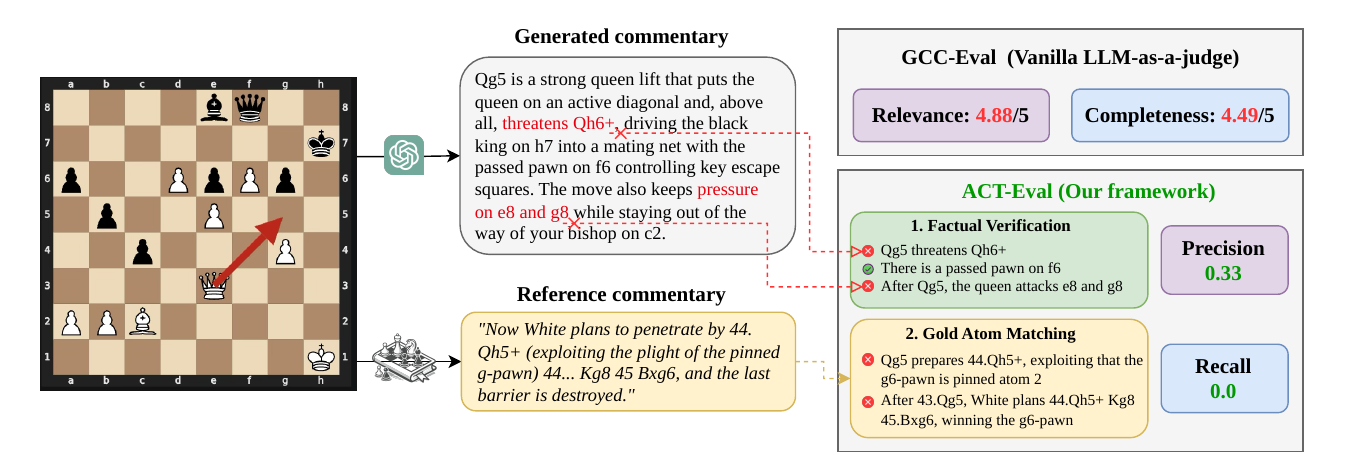}
    \caption{Comparison between GCC-Eval and ACT-Eval on generated chess commentary. The commentary in this example is fluent, but not faithful to the underlying chess position. Nevertheless, GCC-Eval, a naive LLM judge, assigns near-perfect relevance and completeness scores. ACT-Eval instead decomposes the commentary into atomic claims, verifies factual claims using chess tools and engine analysis, and measures coverage against expert reference atoms. Here, it correctly determines that only one of three factual claims is correct, and none of the key expert ideas are covered.}
    \label{fig:comparison}
\end{figure*}

The rapid development of Large Language Models (LLMs) offers a promising direction. LLMs excel at multi-step reasoning across many domains~\cite{wei2022chain,openai2024o1} and can generate reasoning traces in natural language that humans can understand. This creates the possibility to bridge the gap between superhuman expertise and educational explanations. However, substantial challenges remain before LLMs can reliably fulfill this role in domain-specific reasoning tasks. Taking chess as an example, even state-of-the-art reasoning LLMs still perform far below expert human players in chess-specific reasoning, let alone specialized engines~\cite{kolasani2025llm,wen2025chessqa}. Moreover, the spatial structure and long-horizon temporal complexity of chess amplify hallucination issues: models frequently misidentify piece locations, threats or propose illegal move sequences, a pattern we quantify in Section~\ref{sec:error-analysis}, resulting in commentary that often sounds plausible despite containing factual errors (Figure~\ref{fig:comparison}). 
\looseness=-1

These generation problems are compounded by an evaluation problem: there is currently no reliable way to measure how well a language model understands specific domains like chess or how accurately it explains the move. Existing benchmarks primarily assess playing strength or chess understanding through tasks with checkable answers~\cite{kolasani2025llm,wen2025chessqa}, rather than free-form explanations of a move. Standard text generation metrics~\cite{papineni2002bleu,lin2004rouge} only measure surface similarity to reference commentary. Recent LLM-as-a-judge approaches~\cite{liu2023g,kim2025bridging}, which use large language models to assess explanations, introduce a circular problem: the judge is subject to the same domain-specific weaknesses as the model being evaluated \citep{szymanski2025limitations}, making it an unreliable arbiter of domain-specific facts. As Figure~\ref{fig:comparison} illustrates, a hallucinated commentary receives 4.9/5 from GCC-Eval~\cite{kim2025bridging} but 2 out of 3 main claims contain factual errors.
\looseness=-1

In this work, we move toward addressing this gap by proposing an evaluation framework for assessing LLM understanding and interpretability in domain-specific reasoning tasks, using chess as a testbed. A key property of chess is that many factual claims such as move legality, attacks, material balance, or board state can be deterministically checked using chess engines and symbolic tools~\cite{stockfish,fiekas2022pythonchess}, whereas higher-level explanations about plans, coordination, or positional ideas cannot be fully reduced to computation alone. Built on this observation, we propose \textbf{ACT-Eval} (\textbf{A}tomi\textbf{C} and \textbf{T}ool-augmented Evaluation), an atom-grounded, tool-augmented framework for evaluating chess commentary. ACT-Eval decomposes commentary into atomic claims and uses LLMs to route each claim to deterministic tools and check factual correctness. In addition, the LLM judge evaluates strategic and explanatory claims to measure coverage through atomic recall against expert-annotated gold references, capturing whether the explanation identifies the key ideas a strong player would recognize. 
\looseness=-1

Our contributions are summarized as follows:
\begin{itemize}
    \item We introduce ACT-Eval, an evaluation framework that integrates atomic-claim decomposition with tool-augmented verification for chess commentary. By routing computationally decidable claims to engine and board-state oracles,  ACT-Eval grounds the judge in deterministic board-state and engine evidence wherever suitable computational checks are available, reducing reliance on the judge’s parametric chess knowledge.

    \item We construct a benchmark of 325 position-move pairs, covering pedagogical textbooks, modern tournament games, and positions selected for likely human errors. Central to this benchmark is a subset of positions with expert-verified gold atoms derived from expert human annotations. They serve as reference for evaluating conceptual coverage -- whether generated commentary captures the strategic and tactical insights behind a move -- as a complementary dimension to factual correctness.
    \item We evaluate frontier and open-weight LLMs and show that factual hallucinations remain pervasive in chess commentary, particularly in tactical variations and long move sequences. We find that models without engine access cannot reliably distinguish sound moves from mistakes. Tool augmentation substantially improves factual correctness, but atomic recall against expert annotations remains limited. Human validation studies further show that ACT-Eval achieves agreement with expert annotators within the range of inter-human agreement.

\end{itemize}

Using chess as a testbed for domain-specific explanation, we show that current frontier LLMs fail to translate their strong general reasoning into reliable domain-specific expertise, these results suggest that analogous evaluation gaps may arise in other specialized domains. Broadly, ACT-Eval offers a candidate methodological template for domains in which part of the truth is computationally verifiable and the remainder can be anchored in expert annotation. We release the framework and benchmark to support future work on reliable explanation in expert domains.\footnote{\url{https://github.com/hebbarashwin/act_eval}}
\looseness=-1

\section{Related Work}
Reliable evaluation of generated text has been approached from three directions: similarity to human references~\cite{papineni2002bleu,lin2004rouge,zhang2019bertscore}, factual verification against a knowledge source~\cite{min-etal-2023-factscore,chern2025factool,wei2024long}, and LLM-as-a-judge scoring~\cite{liu2023g}. When it comes to domain-specific text generation like chess, each falters for a different reason. Similarity metrics reward fluent text regardless of correctness \citep{kryscinski2020evaluating,maynez2020faithfulness}. Factuality methods verify claims against retrieved evidence or with general-purpose tools such as search and code execution, but verifying a chess claim requires stateful computation against the live position. Verifying a single chess claim may require selecting the correct position, simulating a variation, checking legality and attacks, and composing these with engine evaluation. This is a computation over an evolving board state rather than a lookup, unlike decompose-and-verify frameworks built around retrieval and general-purpose tools \citep{min-etal-2023-factscore,chern2025factool,wei2024long}. LLM judges degrade sharply in domains requiring expert knowledge \citep{szymanski2025limitations}, and a judge without reliable chess knowledge cannot detect a fluent but false chess claim. The closest prior work in our setting, GCC-Eval~\cite{kim2025bridging}, augments an LLM judge with reference commentary and engine analysis but inherits both noisy references and uninterpreted engine scores. ACT-Eval addresses these limitations by routing factual claims to deterministic chess tools and scoring conceptual coverage against expert-verified gold atoms; we provide an extended discussion in Appendix~\ref{sec:related-work}.

\section{The ACT-Eval Framework}
\label{sec:framework}
\begin{figure*}
  \centering
  \includegraphics[width=\linewidth]{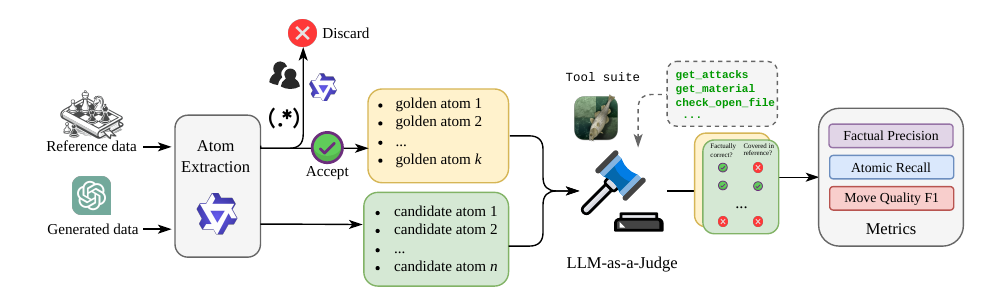}
  \caption{Overview of the ACT-Eval pipeline. Reference commentary is filtered and decomposed into expert-verified gold atoms, while generated commentary is decomposed into candidate atoms. The judge verifies tool-checkable candidate claims using chess-specific board-state and engine tools, matches verified candidates to the gold atoms, and computes factual precision, atomic recall, and move-quality F1. These metrics operationalize the three dimensions introduced in Section~\ref{sec:background}: Factual Precision measures verifiable factual correctness, Atomic Recall measures coverage relative to the gold annotations, and Move-Quality F1 measures detection of engine-labeled mistakes and blunders.}
  \label{fig:eval-pipeline}
\end{figure*}
\subsection{Background} \label{sec:background}
\paragraph{Chess commentary evaluation.} 
We consider the task of evaluating the quality of chess move commentary. Formally, given a board state $s$ represented in a standard encoding (FEN or ASCII) and a move $a$ played in that state, a commentary $\mathcal{C}$ is a natural language explanation that explains the strategic or tactical purpose of the move. Our goal is to assess the quality of $\mathcal{C}$ along three dimensions:
\textit{(1) Factual correctness}, the claims made in $\mathcal{C}$ about the current or subsequent states should be factually correct with respect to the board state. For example, claims about piece placement, attacks, threats, checks, captures, legal moves, and resulting positions should be correct with respect to $s$ and $a$; \textit{(2) Move-quality judgement}, $\mathcal{C}$ should correctly characterize the quality of the move $a$, distinguishing, for example, a strong positional choice from an inaccuracy or blunder; \textit{(3) Conceptual coverage}, $\mathcal{C}$ should identify the key strategic and tactical themes that a human expert would consider most salient in the position.

\paragraph{Why evaluation is hard.} Assessing whether the commentary satisfied these properties is non-trivial. Existing reference-based, LLM-as-a-judge approaches~\cite{kim2025bridging} suffer from two major limitations. First, the reference data the judge relies on are often noisy, containing stylistic artifacts and incomplete analysis. Second, and more critically, LLM judges are insensitive to factual errors: a fluent, chess-specific but \emph{wrong} claim will receive a high score, because the judge lacks the chess knowledge needed to detect the error (see Section~\ref{sec:gcc_main}).

To address these limitations, we propose \textbf{ACT-Eval} (\textbf{A}tomi\textbf{C} and \textbf{T}ool-augmented Evaluation), a framework that combines atom-grounded expert annotations as gold reference and tool-augmented fact verification to substantially reduce LLM hallucination.

\subsection{Gold Atom Extraction} \label{s:atom_extraction}

We collect reference commentary from canonical chess books and annotated online studies. Because source quality varies, we apply regex filters to remove bare move notation, single-word evaluations, and generic filler, then use an LLM with few-shot demonstrations to decompose the remainder into gold atoms: discrete, position-specific factual statements. Each atom is contextual, carrying the full move sequence from the root position so that it remains independently verifiable (Appendix~\ref{app:goldatoms}). Annotations that contradict engine evaluation, and positions offering no concrete positional analysis, are rejected. Human experts review the resulting atoms; Section~\ref{sec:human} reports a validation study finding the gold sets fully complete on 92\% and 88\% of sampled positions.

\subsection{Verification Architecture} \label{sec:verification}

  Our fact-checking pipeline operates in three stages (Figure~\ref{fig:eval-pipeline}):

  \paragraph{Stage 1: Atomic Claim Decomposition}
  The generated commentary is first decomposed into \textbf{atomic claims} (atoms), following the same principles as gold atom extraction (Sec.~\ref{s:atom_extraction}). Each atom is a position-specific factual statement, with full positional context. To enable tool-augmented evaluation, the LLM further breaks down each candidate atom into one or more \textbf{verifiable sub-claims} - specific facts that can be independently verified by tools. For example, the candidate atom ``Qe4 centralizes the queen and attacks the bishop on b4'' decomposes into two sub-claims: (1) ``Qe4 centralizes the queen'', (2)
  ``Qe4 attacks the bishop on b4''. We use a lightweight classifier to assign semantic tags (e.g.\ quality, positional, move continuation) to each sub-claim.
  \looseness=-1

  \paragraph{Stage 2: Tool-Augmented Verification}
  Each atomic claim is verified in a single judge call using a specialized prompt conditioned on the extracted tags and equipped with \textbf{mandatory access to chess tools}. Within this process, the judge evaluates all associated sub-claims before producing a final atom-level verdict. Sub-claims within an atom are verified jointly rather than as independent propositions: the judge receives the original atom and all constituent sub-claims together and must check both their individual truth and any stated relation among them, such as a setup–consequence or causal relation, before issuing the atom-level verdict. By grounding verification in deterministic tool outputs rather than relying on parametric model knowledge, we substantially reduce chess-specific hallucination during factual verification. An atomic claim is marked \textit{correct} only if all constituent sub-claims are verified as true; if any sub-claim is incorrect, the entire candidate atom is marked incorrect.
  \looseness=-1

  \paragraph{Tool Suite}
  We implement a suite of 15+ chess-specific verification tools spanning board-state queries, attack/defense analysis, move simulation, and engine evaluation; the full tool list is provided in Appendix~\ref{sec:tool_verification}.
   Together, these tools enable the judge to ground factual verification in deterministic outputs rather than internal chess knowledge. We demonstrate an example with one atomic claim below:
\looseness=-1

\begin{tcolorbox}[
  title={\textbf{Multi-Tool Verification Example}},
  colback=yellow!5,
  colframe=orange!70!black,
  breakable,
  fonttitle=\bfseries\small,
  fontupper=\small
]
\textbf{Atomic Claim:} Qe4 centralizes the queen and creates a dual threat: it attacks the bishop on b4 and prepares to infiltrate via the e-file. [{\color{red}\texttimes} 1/3 verified]

\begin{itemize}[leftmargin=*, itemsep=2pt, topsep=3pt]
  \item[{\color{green!60!black}$\checkmark$}] centralizes the queen — \texttt{get\_piece\_at}(\texttt{e4}), \texttt{get\_squares}(\texttt{queen}): moved b1$\rightarrow$e4.
  \item[{\color{red}\texttimes}] attacks b4 — \texttt{get\_attacks}(\texttt{e4}) does not include b4.
  \item[{\color{red}\texttimes}] infiltrates the e-file — \texttt{get\_attacks}(\texttt{e4}): file blocked by the e5 pawn.
\end{itemize}
\end{tcolorbox}

  \paragraph{Stage 3: Gold Atom Matching}
  After verification, the judge compares verified atoms from the generated commentary against the gold atoms (Sec.~\ref{s:atom_extraction})
  to measure coverage of important positional concepts.

\subsection{Evaluation Metrics} \label{sec:metrics}

Our evaluation computes three complementary metrics, one per dimension of
Section~\ref{sec:background}.

\begin{itemize}
\item \textbf{Factual Precision:} our proxy for factual correctness.
The fraction of atomic claims verified as factually correct. An atomic claim
is factually correct if and only if all of its sub-claims are verifiably
correct. Claims that are purely strategic or otherwise not verifiable through
tool calls are excluded: the metric therefore measures verifiable factual
correctness rather than factual correctness in general. We also report
\textbf{sub-claim error rate}, the fraction of verifiable sub-claims that fail
tool-based verification.

\item \textbf{Atomic Recall:} our proxy for conceptual coverage.
The fraction of gold atoms (from reference annotations) that are covered by
the generated commentary, computed from Stage 3 matching. 
Coverage is measured relative to our annotation set rather than to all
valid expert analysis, so a commentary that develops a different but sound
line is not credited; any concepts the commentary adds beyond the gold set is not penalized. Appendix~\ref{app:coverage} gives a worked example.

\item \textbf{Move Quality F1:} our proxy for move-quality
judgement. 
The F1 of the commentary's verdict on the played move against an engine-derived quality label, treating mistakes and blunders as the positive class. Accuracy is uninformative here because the classes are unbalanced: 224 of 265 scored moves are sound, so a commentary that never criticises anything scores 84.5\% accuracy and 0 F1. We score only clearly good and
clearly bad moves, excluding the marginal band where the label itself is
unreliable, and we score the verdict rather than the reasoning behind it
(Appendix~\ref{app:quality}).
\end{itemize}

\section{Experiments}

\subsection{Experimental Setup}

\paragraph{Datasets.}

\begin{table}[h]
\centering
\small
\begin{tabular}{lccc}
\toprule
\textbf{Dataset} & \textbf{Size} & \textbf{Source} & \textbf{Gold Atoms} %
\\
\midrule
Textbook & 75 & Chernev (2003) & \checkmark %
\\
Candidates50 & 50 & Lichess-Study & \checkmark %
\\
Critical & 200$^\dagger$ & Lichess+Maia2 & \texttimes %
\\
\midrule
\textbf{Total} & \textbf{325} & & \\
\bottomrule
\end{tabular}
\caption{Evaluation datasets. Textbook and Candidates50 include expert-verified gold atoms and support all three metrics; Critical positions have no reference annotations and evaluate factual correctness and move quality only. $^\dagger$100 positions, each paired with the engine-best move and a Maia2-predicted human mistake.}
\label{tab:datasets}
\end{table}

Table~\ref{tab:datasets} summarizes the three evaluation datasets ($N=325$ total). Textbook and Candidates50 include expert-verified gold atoms and support all three metrics; Critical positions are generated programmatically using Maia2~\citep{tang2024maia} and evaluate only factual correctness and move quality. Dataset construction details are provided in Appendix~\ref{sec:dataset_appendix}.

\paragraph{Models.}
We evaluate six leading proprietary and open-weight models: GPT-5.4, Claude Opus 4.7, Gemini 3.1 Pro, DeepSeek V4 Pro, Qwen3-32B, and Qwen3-8B. The model identifiers are listed in Table~\ref{tab:models}. We evaluate all non-Qwen models both with and without access to chess tools during generation. The Qwen models are evaluated only in the tool-free setting and are run in No-Think mode, with reasoning traces disabled, because preliminary experiments showed that thinking mode produced degenerate outputs (Appendix~\ref{app:qwen}).

\looseness=-1 

\paragraph{Commentary generation.}
Models are prompted to act as a chess instructor explaining a move to an intermediate player, receiving only the board state as a FEN string and the move played. By intermediate we mean club-level players who know standard tactical and positional concepts but benefit from seeing how they apply in a specific position; beginners lack this vocabulary, while expert-level annotation largely replaces prose with variations. Our gold sources are prose explanations aimed at this audience, and our filters remove notation-only annotation (Section~\ref{s:atom_extraction}). In the tool-enabled condition they may invoke the Stockfish-backed tool suite (Table~\ref{tab:tool-suite}) before writing; otherwise they rely entirely on internal chess knowledge. Generation settings and the full prompt are in Appendix~\ref{app:impl}.

\paragraph{Evaluation pipeline.}
All generated commentary is evaluated using the ACT-Eval framework (Section~\ref{sec:framework}) with GPT-5.4 as the judge model, equipped with the full tool suite. Stockfish~14.1  serves as the backend engine for all tool calls, at depth~22 during generation and depth~18 during verification.

\subsection{Main Results}

\begin{table*}[ht]
\centering

\small
\setlength{\tabcolsep}{4.5pt}
\begin{tabular}{llcccc}
\toprule
\textbf{Model} & \textbf{Tools} & \textbf{Error Rate} ($N=325$) & \textbf{Precision} ($N=325$) & \textbf{Recall} ($N=125$) & \textbf{Quality-F1} ($N=265$) \\
\midrule
\multirow{2}{*}{GPT-5.4}
    & \checkmark & 9.2\,{\scriptsize[7.8, 10.6]}  & 59.4\,{\scriptsize[56.3, 62.4]} & 44.0\,{\scriptsize[37.0, 50.9]} & \textbf{97.6}\,{\scriptsize[93.6, 100.0]} \\
    & \texttimes & 22.0\,{\scriptsize[19.9, 24.1]} & 53.9\,{\scriptsize[50.6, 57.2]} & 44.6\,{\scriptsize[37.9, 51.4]} & 12.5\,{\scriptsize[2.7, 22.8]} \\
\midrule
\multirow{2}{*}{Claude Opus 4.7}
    & \checkmark & 13.4\,{\scriptsize[12.1, 14.8]} & 56.4\,{\scriptsize[54.0, 58.7]} & \textbf{58.3}\,{\scriptsize[51.8, 64.6]} & 93.0\,{\scriptsize[86.7, 97.9]} \\
    & \texttimes & 20.8\,{\scriptsize[18.8, 22.8]} & 55.3\,{\scriptsize[52.4, 58.2]} & 45.3\,{\scriptsize[38.5, 51.9]} & 22.2\,{\scriptsize[10.1, 34.3]} \\
\midrule
\multirow{2}{*}{Gemini 3.1 Pro}
    & \checkmark & \textbf{6.7}\,{\scriptsize[5.2, 8.3]} & 52.0\,{\scriptsize[47.8, 56.1]} & 48.2\,{\scriptsize[41.2, 55.5]} & 64.7\,{\scriptsize[50.0, 76.7]} \\
    & \texttimes & 10.0\,{\scriptsize[8.4, 11.7]} & \textbf{74.2}\,{\scriptsize[71.4, 77.0]} & 58.2\,{\scriptsize[51.6, 64.7]} & 25.4\,{\scriptsize[11.9, 38.2]} \\
\midrule
\multirow{2}{*}{DeepSeek V4 Pro}
    & \checkmark & 14.6\,{\scriptsize[13.0, 16.2]} & 54.8\,{\scriptsize[52.0, 57.6]} & 53.2\,{\scriptsize[46.2, 60.2]} & 83.9\,{\scriptsize[75.0, 91.4]} \\
    & \texttimes & 36.7\,{\scriptsize[33.3, 40.1]} & 40.5\,{\scriptsize[37.0, 44.1]} & 30.1\,{\scriptsize[23.9, 36.6]} & 23.9\,{\scriptsize[14.4, 33.0]} \\
\midrule
Qwen3-8B (NT)
    & \texttimes & 55.5\,{\scriptsize[52.2, 58.7]} & 22.0\,{\scriptsize[19.0, 25.2]} & 13.9\,{\scriptsize[9.3, 18.9]} & 27.6\,{\scriptsize[17.1, 37.8]} \\
\midrule
Qwen3-32B (NT)
    & \texttimes & 44.1\,{\scriptsize[40.7, 47.4]} & 24.7\,{\scriptsize[22.0, 27.5]} & 16.0\,{\scriptsize[11.8, 20.6]} & 40.9\,{\scriptsize[29.7, 51.0]} \\
\bottomrule
\end{tabular}
\caption{\textbf{Combined ACT-Eval Results} across all datasets. \textit{Error Rate} is the fraction of sub-claims that fail verification; \textit{Precision} is the fraction of atoms that are factually correct; \textit{Recall} is the fraction of gold reference atoms covered, computed only on datasets with gold reference annotations; \textit{Quality-F1} is the F1 score for detecting bad (mistake/blunder) moves, on positions excluding inaccuracies. Bracketed ranges are $95\%$ bootstrap confidence intervals ($B{=}10{,}000$ resamples, clustered at the position level).  NT = No-Think mode. With tools, Gemini 3.1 Pro and DeepSeek V4 Pro fail to produce any commentary on 29.8\% and 4.0\% of positions, respectively, by exhausting their tool-call budget; these failures are scored as commentary with no claims (zero Precision and Recall, and no verdict on the move).
}
\label{tab:main-results}
\end{table*}

Table~\ref{tab:main-results} presents results aggregated across all three datasets. Tool access substantially reduces factual errors and substantially improves move-quality
judgement; its effect on conceptual coverage (recall) is model-dependent, and coverage remains limited in every configuration.

\paragraph{LLMs hallucinate frequently in chess commentary.}
Without tool access, even frontier models produce factually unreliable commentary: sub-claim error rates run from $10.0\%$ (Gemini 3.1 Pro)
to $55.5\%$ (Qwen3-8B).
These results are consistent with the well-documented finding that LLMs lack reliable chess-playing ability~\cite{kolasani2025llm,toshniwal2022chess,wen2025chessqa}: generating good commentary requires not just fluency but the ability to identify key ideas in a position, which in turn demands accurate board comprehension - recognizing piece locations, legal moves, tactical patterns, and threats.
Errors in this basic spatial reasoning propagate directly into the explanations.
\looseness=-1

\paragraph{Tool access substantially reduces factual errors.}
Equipping models with chess tools, including Stockfish evaluation and variations, legality checks, piece location queries, and attack/defense analysis - mitigates a large share of these errors.
GPT-5.4's sub-claim error rate drops from 22.0\% to 9.2\% with tool access, more than halving the rate of verification failures, and cuts
Claude Opus' by a third. Similar trends are observed on Gemini 3.1 Pro and DeepSeek V4 Pro.

Tools shift the burden: models need not compute internally whether a square is attacked or a move is legal, and can concentrate on structuring an explanation around verified facts. They do not remove the need for chess understanding: the model must still select which facts matter, interpret what the tools return, and assemble them into an explanation, and the coverage results below show that this remains hard.
\looseness=-1

\paragraph{Recall remains low across all models.}
Even the best configurations, Claude Opus 4.7 with tools and Gemini 3.1 Pro without tools, cover only 58.3\% and 58.2\% of expert-identified gold atoms, and the Qwen models reach at most 16\%. The effect of tools is model-dependent: recall is essentially unchanged for GPT-5.4 (44.6\% $\to$ 44.0\%) and rises sharply for Claude Opus 4.7 (45.3\% $\to$ 58.3\%) and DeepSeek V4 Pro (30.1\% $\to$ 53.2\%). Gemini's recall falls with tools (58.2\% $\to$ 48.2\%) because it produces no commentary on 25 of the 125 gold positions after exhausting its tool-call budget; on the remaining positions its recall is 60.3\%. Verified facts are therefore not on their own sufficient to surface expert ideas; whether a model builds on them varies by model. Low recall is not a ceiling imposed by the gold set: taking the best configuration on each position raises recall to 83.1\%, and only 4 of 125 positions contain no gold atom that any configuration recovers. Each model omits a different subset of the expert ideas. Identifying the critical ideas in a position remains the hardest of the three dimensions.

\begin{revblock}\revcol
\paragraph{Models detect bad moves unreliably without engine access.}
Judging move quality is an essential component of a good explanation. Without engine access, bad-move detection is weak across all evaluated models: treating mistakes and blunders as the positive class, F1 ranges from 12.5 for GPT-5.4 to 40.9 for Qwen3-32B. Tool access raises F1 to 97.6 for GPT-5.4 and 93.0 for Claude Opus 4.7. Because the metric compares commentary stance against engine-derived labels, these gains primarily demonstrate that models can use engine evidence to calibrate their move-quality statements, rather than an improvement in chess evaluation ability. Appendix~\ref{app:quality-pr} provides the per-class results. Gemini 3.1 Pro's lower F1 with tools reflects generation failures rather than misjudgement: all 19 of its missed bad moves are positions where it produced no commentary.
\end{revblock}
\looseness=-1

\begin{revblock}\revcol
\subsection{Comparison with vanilla LLM-as-a-judge}
\label{sec:gcc_main}
To test whether the failure illustrated in Figure~\ref{fig:comparison} is
systematic rather than anecdotal, we score all 1,948 commentaries from six generation configurations (GPT-5.4 and Claude Opus 4.7, each with and without tools; Qwen3-8B; Qwen3-32B) with both GCC-Eval~\citep{kim2025bridging} and ACT-Eval.
ACT-Eval flags 451 of them as majority-erroneous (sub-claim error rate
$\geq 0.5$). On exactly these commentaries, GCC-Eval still awards high
relevance ($\geq 4/5$) to $33.5\%$ and high completeness to $14.0\%$.
Within each configuration, this rate matches the rate over all of that configuration's commentary: $96.9\%$ vs.\ $96.9\%$ for GPT-5.4 without tools, $92.3\%$ vs.\ $93.2\%$ for Claude Opus 4.7 without tools, and $8.0\%$ vs.\ $7.4\%$ for Qwen3-8B. GCC-Eval separates models, largely on fluency, but does not separate correct from incorrect commentary by the same model.
This confirms that the failure mode illustrated in Figure~\ref{fig:comparison} occurs systematically rather than in isolated cases. ACT-Eval addresses it by replacing both subjective dimensions with verifiable ones: relevance with factual precision, computed over atomic claims each checked against the board, and completeness with atomic recall against expert-verified gold atoms. We develop this comparison further in Appendix~\ref{app:gcc}.
\end{revblock}
\looseness=-1

\section{Validation against Human Judgments}
\label{sec:human}

To validate that the ACT-Eval judges align with expert human assessment, we conduct two validation studies targeting the two judge-dependent metrics: (i) factual correctness (Stage 2) and (ii) atomic recall against gold references (Stage 3).

\paragraph{Participants.} Four chess players with Lichess blitz Elo between 1800-2200 participated in the study. All four completed Study 1, while two completed Study 2. Annotators worked through a purpose-built interface that presented the board diagram, FEN, and engine analysis alongside the LLM commentary or claim under review, and they were blinded to the judge's verdicts.
\looseness=-1

\subsection{Study 1: Claim Verification}

We assess whether the judge's per-atom verdict agrees with expert human judgement. We draw 50 atomic claims from the evaluation corpus using stratified sampling across different models and judge verdicts (30\% judge-correct, 70\% judge-incorrect; fixed seed) and place them in randomized order. For each item, annotators (a) issued a verdict on a four-point scale (Correct / Mostly correct / Incorrect / Cannot determine)
and (b) reported confidence (low/medium/high). 
\looseness=-1

Since the judge produces binary labels and marks an atom incorrect whenever any sub-claim fails verification (Section~\ref{sec:verification}), mostly correct human verdicts are mapped to incorrect for agreement analysis, consistent with the judge's own aggregation rule.
\looseness=-1

\paragraph{Judge agreement approaches inter-human variability.} We measure agreement between the judge's binary verdict and human annotators on the 50 sampled claims. Per-annotator agreement with the judge ranges from 69\% to 83\%, with a macro-average of 74.5\% across the four annotators. Pairwise agreement between annotators ranges from 73\% to 91\%. This indicates that the judge falls within the range of inter-human agreement on the task.
\looseness=-1

Pairwise Cohen’s $\kappa$ ranges from 0.107 to 0.697, while Fleiss’ $\kappa$ is 0.285. Because the sample is 70\% judge-incorrect by design, these chance-corrected measures are prevalence-sensitive and should be interpreted alongside raw agreement.
\looseness=-1

\paragraph{The judge is  reliable at detecting errors, but misses some.} When the judge marks a claim \emph{incorrect} ($n{=}35$), annotators agree 83--94\% of the time: errors flagged by the judge are almost always real errors. In 21 of these 35 rejections exactly one sub-claim failed, so agreement here directly confirms a sub-claim verdict. When it marks a claim \emph{correct}, agreement drops to 14--57\%. The judge therefore catches many genuine errors but also misses errors identified by human annotators. ACT-Eval is thus validated chiefly as an error detector; because the judge-correct sample is small, the reported error rates in Table~\ref{tab:main-results} should be read as judge-dependent estimates rather than exact counts.
\looseness=-1

\subsection{Study 2: Commentary Coverage}
\label{s1_results}

We evaluate whether the judge's recall metric aligns with expert human assessment over 25 position--move pairs. The task proceeds in two phases. In the \emph{completeness check}, annotators are shown the gold atoms and judge whether they cover the most important observations about the move (fully / partially / no). Annotators can optionally add missing key points, which are appended to the gold set. This provides a sanity check on the quality of our extracted gold atoms. 
\looseness=-1

Next, in the \emph{coverage annotation}, annotators are shown an LLM-generated commentary (sampled across models) and indicate, for each gold atom (including any additions), whether it is covered by the commentary. Human recall is computed as the fraction of (augmented) gold atoms judged covered.
\looseness=-1

\paragraph{Gold atoms serve as adequate coverage targets.} Gold atoms are extracted from expert commentary, filtered, and decomposed via the pipeline of Section~\ref{s:atom_extraction}. Across the 25 sampled positions, the two annotators rated the gold set as fully complete in 92\% and 88\% of cases respectively; the remaining items were marked partially complete, typically with a single additional point appended. This validates the extracted gold atoms as adequate coverage targets.
\looseness=-1

\paragraph{Judge recall correlates strongly with expert humans.} The judge's per-position recall correlates with the two annotators at Pearson $r=0.732$ and $r=0.749$. The annotators themselves agree at $r=0.835$, indicating that the judge-human gap is small relative to the inherent variability of the task.
\looseness=-1

The judge applies a stricter coverage threshold than human experts: mean judge recall over the sample is 0.33, compared to 0.41--0.47 for the two annotators, with the gap consistent across positions. Absolute recall values in Table~\ref{tab:main-results} are therefore judge-dependent and lower than human annotators would assign on this sample, while the strong correlation supports comparisons between configurations.
\looseness=-1
\section{Analysis}
\subsection{Error Analysis}
\label{sec:error-analysis}

\begin{figure}
  \centering
  \includegraphics[width=\linewidth]{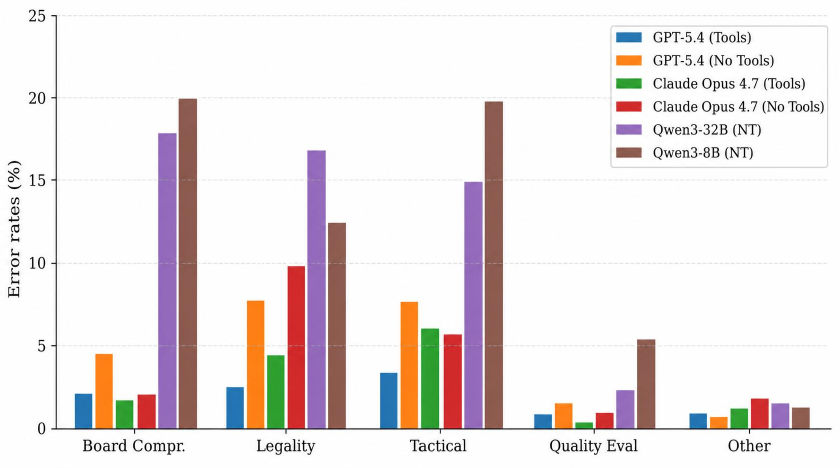}
  \caption{Factual error breakdown by category across models, normalized by 
total decisive sub-claims.}
  \label{fig:error-rates}
\end{figure}

Figure~\ref{fig:error-rates} breaks down factual errors across five categories (Appendix~\ref{app:error_taxonomy}), normalized by decisive sub-claims. Open-weight models are dominated by board comprehension and legality failures: pieces hallucinated on wrong squares, misidentified attackers, illegal continuations, this indicates weak internal representations of the board state itself.
Frontier models largely avoid these static failures but still err in multi-move variations without tool access: GPT-5.4 and Claude Opus 4.7 generate plausible continuations that become inconsistent after several moves, showing that linguistic fluency does not extend to maintaining board state over long-horizon sequences.
Tool augmentation shifts the distribution rather than uniformly shrinking it. Legality checks and board-state queries largely eliminate low-level hallucinations, leaving errors concentrated in tactical reasoning: models misjudge whether an idea works, overlook defensive resources, or hallucinate forcing continuations. Errors in the quality-evaluation category become rare, consistent with the Move Quality F1 gains in Table~\ref{tab:main-results}.

\subsection{Ablation Study on Tool Augmentation}

To isolate the contribution of tools to the judge, we hold decomposition fixed and re-verify the same sub-claims with GPT-5.4 without tool access. On 29 claims from Study 1 that all four annotators marked incorrect with high confidence (a clean set of confirmed errors), ACT-Eval misses 21\%, against 48\% for the no-tool ablation. Tool augmentation therefore accounts for most of the judge's error-detection ability, and the residual 21\% shows that the judge still misses some confirmed errors.

\section{Conclusion}
We presented ACT-Eval, a framework that evaluates chess commentary by decomposing it into atomic claims, verifying each claim with chess engines and board-state tools, and matching it against expert-verified gold atoms. This separates three properties that single-score judges conflate: verifiable factual correctness, coverage of expert ideas, and move-quality judgement. On a benchmark of 325 position--move pairs, we find that LLMs hallucinate frequently when explaining chess: without tools, sub-claim error rates range from 10.0\% to 55.5\%. Tool access substantially reduces factual errors and substantially improves move-quality judgement, but coverage of expert ideas remains limited for every model, so access to correct evidence does not by itself produce a good explanation. Validation against expert annotators shows that the judge's verdicts fall within the range of inter-human agreement and that its recall correlates strongly with human coverage judgements, while also revealing residual errors. Selecting which ideas to explain, and verifying the strategic claims that tools cannot check, remain the main open challenges for chess commentary and, more broadly, for explanation in domains where only part of the truth is computable.

\clearpage
\section*{Limitations}
\paragraph{Tool-suite coverage.} Our verification framework relies on a fixed set of chess-specific tools. Human validation shows that ACT-Eval still misses some factual errors. These missed errors are not confined to a single category; they occur across board comprehension, legality, tactical reasoning, and evaluation-related claims. Because ACT-Eval can both miss factual errors and reject valid claims, its reported error rates are judge-dependent estimates rather than formal lower bounds. Human validation nevertheless shows that some expert-identified errors are missed. Expanding the suite, particularly for long-horizon variation checking and endgame-specific patterns, is a natural direction for closing this gap. 

\paragraph{The framework still inherits LLM-judge limitations.} Although tools reduce reliance on the judge's internal chess knowledge, the judge still interprets claims, selects tools and tool-arguments, and aggregates evidence. Appendix~\ref{app:residual} documents failures in both directions on the same claims: errors the GPT-5.4 judge accepts that an independent Gemini judge catches, and valid claims it rejects by misreading tool output. The deterministic tools returned correct evidence in both cases; the verification procedure around them failed.

\paragraph{LLM-based decomposition can introduce errors.} ACT-Eval relies on an LLM to split commentary into atomic claims. This step can produce incorrect or under-contextualized atoms, especially when claims depend on earlier sentences, implicit move sequences, or surrounding tactical context. If an atom loses necessary context, the verification stage may incorrectly reject it.
\paragraph{Gold atoms are incomplete approximations of expert understanding.} The expert-verified gold atoms capture many salient tactical and strategic ideas, but chess explanations are open-ended and multiple valid analyses may exist for the same move. Atomic recall should therefore be interpreted as coverage relative to our annotation set, not as an absolute measure of all possible expert insight. Recall also weights all gold atoms equally, although the importance of a point depends on the reader's level; it is not a measure of pedagogical usefulness, which we do not evaluate with target readers.

\paragraph{Evaluation scale.} The benchmark contains 325 position–move pairs, with gold atoms for 125, while the human validation studies cover 50 atomic claims and 25 commentaries. These samples support calibrated comparisons among the evaluated systems but do not establish performance across the full diversity of chess positions, commentary styles, or annotators.

\paragraph{Evidence is chess-specific.} We frame ACT-Eval as a template for domains where part of the truth is computable and part requires expert judgement, but we validate it only on chess. Transfer requires both a deterministic oracle for the verifiable component and expert annotation for the rest; whether those conditions are commonly met is left to future work.

\section*{Acknowledgments}
We thank Amey Bhat, Praveen Iyer, Sambhu Karumanchi, and Sanskar Jaiswal for their expert chess annotations. We are grateful to Vidit Gujrathi for insightful conversations. Finally, we thank Sentient Labs for generously providing the compute resources used in this work.
\newpage

\bibliography{custom}
\appendix
\newpage
\section{Metric Definitions}

\begin{revblock}\revcol
\subsection{Move Quality}
\label{app:quality}
 
Move Quality F1 compares an engine-derived label for the played move
against the stance the commentary takes toward it.
 
\paragraph{Deriving move quality from the engine.}
Chess engines such as Stockfish score a position in \emph{centipawns}, where
$100$ centipawns is roughly the value of one pawn. A move's centipawn loss --
how much worse the position becomes relative to the engine's preferred move --
cannot be thresholded directly, because the same loss carries different
consequences depending on the position. Giving up $100$ centipawns in a level
position can convert a draw into a loss, whereas the same concession in a
position that is already overwhelmingly won changes nothing about the likely
outcome. Following the Lichess accuracy metric~\citep{lichess-accuracy}, we therefore first map the
evaluation $c$ (in centipawns, from the perspective of the side to move) to an
estimated win probability,
\[
\mathrm{Win\%} = 50 + 50\left(\tfrac{2}{1+e^{-0.00368208\, c}} - 1\right),
\]
and define $\mathrm{wp\_loss} = \mathrm{Win\%}(s,a^{*}) - \mathrm{Win\%}(s,a)$
for played move $a$ and engine best move $a^{*}$. We bin $\mathrm{wp\_loss}$
using the thresholds Lichess applies when annotating games,\footnote{\texttt{Advice.scala} in the Lichess source code, commit \texttt{cd92167}: \url{https://github.com/lichess-org/lila/blob/cd921677f6083fbc3bc4c2c2521544e4b727894f/modules/tree/src/main/Advice.scala}}
so our labels match what a standard public analysis tool reports:
\emph{good} ($\leq 10$), \emph{inaccuracy} ($10$--$20$),
\emph{mistake} ($20$--$30$), \emph{blunder} ($> 30$).
 
\paragraph{Commentary stance.}
Decomposition extracts the stance toward the played move as \emph{good},
\emph{bad}, or \emph{inconclusive} (no explicit judgement), resolving
negations so that ``not the best'' maps to \emph{good}. Judgements about
alternatives or about the position as a whole are excluded.
 
\paragraph{Matching.}
We score commentary on clearly good and clearly bad moves only, excluding
the $60$ of $325$ positions whose engine label falls in the \emph{inaccuracy}
band ($10 < \mathrm{wp\_loss} \leq 20$), leaving $n{=}265$. Nearly all of these
excluded positions ($59$ of $60$) lie in the Critical set, which by design
samples positions where Maia2 predicts an inaccuracy or mistake; the remaining one is in
Candidates50, and none are in Textbook. Human annotators
routinely disagree with the engine at this margin, often for sound practical
reasons; scoring stance there would measure agreement with a cutoff rather
than whether a model can tell a good move from a bad one. On the remaining
positions, a \emph{good} label is satisfied by a \emph{good} or
\emph{inconclusive} stance, since commentary that explains a sound move
without editorialising is not in error, while \emph{mistake} and
\emph{blunder} require an explicit \emph{bad} stance, since failing to flag a
losing move is a substantive failure.

\begin{table*}[ht]
\centering

\small
\begin{tabular}{llccc}
\toprule
\textbf{Model} & \textbf{Tools} & \textbf{Precision} & \textbf{Recall} & \textbf{Quality-F1} \\
\midrule
\multirow{2}{*}{GPT-5.4}
    & \checkmark & \textbf{95.3}\,{\scriptsize[88.0, 100.0]} & \textbf{100.0}\,{\scriptsize[100.0, 100.0]} & \textbf{97.6}\,{\scriptsize[93.6, 100.0]} \\
    & \texttimes & 12.8\,{\scriptsize[2.8, 24.3]} & 12.2\,{\scriptsize[2.9, 23.3]} & 12.5\,{\scriptsize[2.7, 22.8]} \\
\midrule
\multirow{2}{*}{Claude Opus 4.7}
    & \checkmark & 88.9\,{\scriptsize[78.8, 97.6]} & 97.6\,{\scriptsize[91.7, 100.0]} & 93.0\,{\scriptsize[86.7, 97.9]} \\
    & \texttimes & 22.5\,{\scriptsize[9.8, 36.1]} & 22.0\,{\scriptsize[9.8, 35.0]} & 22.2\,{\scriptsize[10.1, 34.3]} \\
\midrule
\multirow{2}{*}{Gemini 3.1 Pro}
    & \checkmark & 81.5\,{\scriptsize[65.4, 95.7]} & 53.7\,{\scriptsize[38.1, 69.0]} & 64.7\,{\scriptsize[50.0, 76.7]} \\
    & \texttimes & 30.0\,{\scriptsize[14.7, 45.8]} & 22.0\,{\scriptsize[9.8, 35.5]} & 25.4\,{\scriptsize[11.9, 38.2]} \\
\midrule
\multirow{2}{*}{DeepSeek V4 Pro}
    & \checkmark & 75.0\,{\scriptsize[62.7, 86.5]} & 95.1\,{\scriptsize[87.8, 100.0]} & 83.9\,{\scriptsize[75.0, 91.4]} \\
    & \texttimes & 14.3\,{\scriptsize[8.1, 20.6]} & 73.7\,{\scriptsize[52.6, 92.9]} & 23.9\,{\scriptsize[14.4, 33.0]} \\
\midrule
Qwen3-8B (NT)
    & \texttimes & 20.7\,{\scriptsize[12.2, 30.0]} & 41.5\,{\scriptsize[26.7, 56.7]} & 27.6\,{\scriptsize[17.1, 37.8]} \\
\midrule
Qwen3-32B (NT)
    & \texttimes & 29.7\,{\scriptsize[20.5, 39.3]} & 65.9\,{\scriptsize[51.0, 80.4]} & 40.9\,{\scriptsize[29.7, 51.0]} \\
\bottomrule
\end{tabular}
\caption{\textbf{Move quality assessment.} Treating move-quality assessment as binary detection of \emph{bad} moves (mistake/blunder vs.\ good), on positions excluding inaccuracies. A \emph{bad} prediction requires an explicit \emph{bad} stance; \emph{good} and \emph{inconclusive} both count as not-bad, so silence on a losing move is a miss. Bracketed ranges are $95\%$ position-clustered bootstrap CIs ($B{=}10{,}000$). NT = No-Think mode.}
\label{tab:quality-pr}
\end{table*}

\subsection{Move Quality: Per-Class Results}
\label{app:quality-pr}

Table~\ref{tab:quality-pr} decomposes $\text{F1}_\text{bad}$ into
\emph{Quality-Precision} (of the moves a commentary called bad, how many were
bad) and \emph{Quality-Recall} (of the actually bad moves, how many it caught).
We use the hyphenated names throughout to distinguish these from Factual
Precision and Atomic Recall, which measure unrelated quantities.

The no-tool failures take two distinct forms. GPT-5.4 and Claude Opus 4.7 flag
close to the right number of moves ($39$ and $40$, against $41$ actual) but
largely the wrong ones, with Quality-Precision and Quality-Recall both between $12\%$ and $23\%$ --- the signature of guessing at roughly the correct rate. The Qwen
models instead over-flag, calling $82$ and $91$ of $265$ moves bad; this buys
Quality-Recall ($41.5$ and $65.9$) while Quality-Precision collapses, and the
false alarms on sound moves are what push their accuracy below the $84.5\%$
majority-class baseline. Tool-augmented frontier models are calibrated in both
senses: GPT-5.4 predicts $43$ bad moves and Claude $45$, against $41$.

The positive class has small support ($41$ positions), so intervals are wide
and the robust claim is the tool-versus-no-tool contrast rather than fine
rankings among models. GPT-5.4 with tools has a degenerate Quality-Recall
interval because it caught every bad move in every bootstrap resample.

\end{revblock}

\begin{revblock}\revcol
\subsection{Atomic Recall}
\label{app:coverage}

Atomic Recall is a proxy for conceptual coverage: it measures how many of our
annotated expert observations a commentary mentions. Three consequences follow.

First, coverage is bounded by the annotation. Gold atoms are extracted from a
single expert source per position, and chess positions admit multiple sound
analyses; a commentary that pursues a different but valid line receives no
credit for it. Recall is therefore coverage of \emph{these} expert
observations, not of all expert insight available in the position.

Second, recall does not penalize what a commentary adds. A commentary can
match every gold atom while surrounding them with generic or irrelevant
material. Factual Precision constrains whether the additions are true, but
neither metric asks whether they were worth saying.

Third, matching is performed by an LLM judge. Our validation study
(Section~\ref{sec:human}) finds that the judge applies a stricter
threshold than human annotators ($0.33$ against $0.41$--$0.47$), so reported
recall is lower than human-assessed coverage on that sample.
\end{revblock}

\begin{revblock}\revcol
 
We now provide an example: the following pair illustrates the
distinction between commentaries high and low recall: both commentaries pass factual verification, but only one
recovers the point of the move.
 
\begin{quote}
\small
\textbf{Position:} \texttt{8\fs pn2Q3\fs 1p1p2pk\fs 2pP2rp\fs P1P2q2\fs 1NP5\fs 6B1\fs 2R4K b - - 15 47}\\
\textbf{Move:} \texttt{Qh4+}\\[2pt]
\textbf{Gold atoms:}
\begin{enumerate}[nosep,leftmargin=*]
\item \texttt{Qh4+} forces White's king to move to \texttt{g1}.
\item After \texttt{Qh4+} and \texttt{48.Kg1}, Black can win the queen by \texttt{Rxg2+}.
\end{enumerate}
\end{quote}
 
\paragraph{High coverage (recall = 1.0).}
\begin{quote}
\small
``\texttt{Qh4+} is a forcing move that sets up a brilliant discovered attack
to win the white queen. After the forced \texttt{Kg1}, Black will play
\texttt{Rxg2+}, sacrificing the rook with check. This clears the \texttt{g5}
square, opening the \texttt{h4}--\texttt{e7} diagonal for the black queen to
capture White's undefended queen on \texttt{e7}.''
\end{quote}
Both gold atoms are matched: the commentary names the forced king move and the
\texttt{Rxg2+} resource that wins the queen.
 
\paragraph{Low coverage (recall = 0.0).}
\begin{quote}
\small
``\texttt{Qh4+} is the strongest move because it drags Black's queen from
\texttt{f4} to a square where White cannot challenge it, while forcing the
white king to react. From \texttt{h4} the queen immediately eyes \texttt{h2},
and with the rook already controlling the \texttt{g}-file, Black's attack
becomes decisive: after a natural move like \texttt{Bh3}, Black has
\texttt{Qxh3} mate, and after \texttt{Kg1} Black keeps a huge winning attack
with threats like \texttt{Qh2+}. It's a precise move that turns Black's active
queen and rook into a direct mating net against the exposed king.''
\end{quote}
This commentary is fluent, its individual claims are factually defensible, and
its overall verdict on the move is correct. It nonetheless matches neither gold
atom: it does not identify \texttt{Kg1} as forced, and it never finds
\texttt{Rxg2+}, the concrete resource that decides the game.
\end{revblock}

\section{Framework Details}

\subsection{Gold Atom Extraction} \label{app:goldatoms}
\begin{tcolorbox}[title={\textbf{Golden Atom Extraction Example}}, colback=green!3, colframe=green!50!black, breakable, fonttitle=\bfseries\small, fontupper=\small]
  \textbf{Original annotation:} ``Nxd7 removes one of the defenders of the knight on f6, which is a typical tactical theme in such positions. After the expected Nh5, White
   can capture the bishop on g6, opening up Black's king. Controlling the center remains paramount in the middlegame.''
 
  \medskip        
  \textbf{\textcolor{red}{\texttimes} BAD (non-contextual — later atoms lose move context):}
  \begin{itemize}
      \item Nxd7 removes one of the defenders of the knight on f6.
      \item After Nh5, White can capture on g6. \textcolor{red}{← Missing context: after what?}
      \item Capturing on g6 opens Black's king. \textcolor{red}{← After which moves?}
      \item \sout{``which is a typical tactical theme in such positions''} — generic, not position-specific
      \item \sout{``Controlling the center remains paramount in the middlegame''} — not relevant to this move
  \end{itemize}

  \medskip
  \textbf{\textcolor{green}{\checkmark} GOOD (contextual, standardized):}
  \begin{itemize}
      \item Nxd7 removes one of the defenders of the knight on f6.
      \item After Nxd7 Nh5, White can capture the bishop on g6.
      \item After Nxd7 Nh5, capturing on g6 opens up Black's king to attack.
  \end{itemize}
  \end{tcolorbox}

\begin{table*}[ht]
\centering

\footnotesize
\vspace{2pt}
\begin{tabular}{lll}
\toprule[1pt]
\textbf{Category} & \textbf{Tool} & \textbf{Purpose} \\
\midrule[0.75pt]

Position Query 
& \texttt{get\_piece\_at}(fen, sq) 
& Return the piece on a given square \\

& \texttt{get\_legal\_moves}(fen) 
& List all legal moves in the position \\

& \texttt{is\_check}(fen) 
& Determine whether the side to move is in check \\

& \texttt{get\_material}(fen) 
& Compute material balance / piece counts \\

& \texttt{get\_squares}(fen, piece, color) 
& Find locations of a piece type \\

\midrule[0.25pt]

Attack / Defense 
& \texttt{get\_attacks}(fen, sq) 
& List squares attacked by a piece on a square \\

& \texttt{get\_attackers}(fen, sq) 
& List pieces attacking a square \\

& \texttt{count\_attackers\_defenders}(fen, sq) 
& Count attacking and defending pieces on a square \\

& \texttt{is\_pinned}(fen, sq) 
& Check whether a piece is pinned \\

& \texttt{check\_threat}(fen, move) 
& Verify existence of a tactical or strategic threat \\

\midrule[0.25pt]

Geometry 
& \texttt{check\_ray\_alignment}(sq\_a, sq\_b) 
& Test alignment along ranks, files, or diagonals \\

& \texttt{check\_open\_file}(fen, file, color) 
& Check if a file is half-open \\

\midrule[0.25pt]

Variations 
& \texttt{try\_variation}(fen, moves) 
& Simulate a move sequence  \\

\midrule[0.25pt]

Engine Analysis 
& \texttt{eval\_move}(fen, move) 
& Evaluate move quality (e.g., win probability loss) \\

& \texttt{get\_engine\_eval}(fen, move) 
& Get Stockfish evaluation of the move and best move \\

& \texttt{get\_top\_moves}(fen, n) 
& Return top-$n$ engine-recommended moves \\

& \texttt{compare\_moves}(fen, m\_a, m\_b) 
& Compare relative quality of two moves \\

\bottomrule[1pt]
\end{tabular}
\caption{\textbf{Tool suite.} We implement a set of tools that enables the LLM judge to verify the factual correctness of diverse chess claims through chained tool calls.}
\label{tab:tool-suite}
\end{table*}

\subsection{Tool-augmented Verification} \label{sec:tool_verification}

Our verification framework exposes a collection of chess-specific tools designed to support fine-grained factual checking of generated commentary. The tools operate over FEN-encoded positions and return structured outputs that can be chained together by the judge model during verification. We group the tools into five categories: (i) board-state and legality queries, (ii) attack and defense queries, (iii) geometric reasoning, (iv) move-variation simulation, and (v) engine analysis queries.

Importantly, many commentary claims require composing multiple tools rather than relying on a single engine score. For example, verifying a claim about a tactical threat may involve checking move legality, simulating a continuation, evaluating resulting attacks, and comparing engine assessments before and after the sequence. This modular design allows ACT-Eval to verify diverse classes of chess claims while minimizing reliance on the judge model’s internal chess knowledge.

Claims that a move is merely available (e.g., a threat or a plan) are checked for legality in the position reached by the stated line, whereas claims about outcomes (winning material, preventing a move, or a move being forced) are checked with engine evaluation of the line and its alternatives. The full rules are given in the verification prompt (Appendix~\ref{app:prompts}).

\subsection{Error Taxonomy} \label{app:error_taxonomy}
We define five error categories used in Figure~\ref{fig:error-rates}:

\begin{itemize}
    \item \textbf{Board comprehension}: Failures to accurately represent the static 
    board configuration, including piece identity, location, and structural properties 
    (e.g., open vs.\ blocked files), reflecting an incorrect or incomplete internal 
    representation of the position.

    \item \textbf{Legality}: Failures to track board state across moves, resulting in 
    proposed moves or sequences that are not legally playable from the given position.

    \item \textbf{Quality evaluation}: Incorrect assessments of move strength or 
    positional advantage that contradict engine analysis, including invalid positional 
    comparisons (e.g., claiming a feature changes when it does not).

    \item \textbf{Tactical reasoning}: Errors in reasoning about the dynamic 
    consequences of a position, such as asserting nonexistent threats or claiming a move is forced despite viable alternatives, often downstream of board 
    comprehension failures.
    \item \textbf{Other}: Residual category for all other errors.
\end{itemize}

\subsection{Prompts}
\label{app:prompts}
\begin{promptbox}[Decomposition prompt]
Extract atomic factual claims from a chess move explanation.

CRITICAL: ONE CLAIM PER ATOM. Each atom must contain exactly ONE verifiable fact. \
If a sentence contains multiple claims, split it into separate atoms.

CRITICAL: Extract ONLY positional/tactical facts about the move or resulting position on the board. \
Drop historical facts (dates, tournaments, "X defeated Y"), attributions ("Gunsberg wrote...", \
"Lasker called..."), player-choice biography ("Lasker evaded this line"), and opening-popularity \
or reception claims — these are not board-verifiable.

EXAMPLE — given this text:
"Bf3 allows Black to play Bf5, skewering the queen on b1 and the knight on c3."

BAD (compound claims):
- "Bf3 allows Black to play Bf5, skewering the queen on b1 and the knight on c3."

GOOD (atomic claims - one fact each):
- "Bf3 would allow Black to play Bf5"
- "After Bf3 Bf5, the white queen is on b1"
- "After Bf3 Bf5, there is a white knight on c3"
- "After Bf3 Bf5, the bishop on f5 skewers the queen on b1 and knight on c3"

CONTEXTUAL ATOMS: Each atom must be self-contained with full positional context. \
If an atom references a position after a sequence of moves, include the FULL \
preceding move sequence.

EXAMPLE — given this text:
"Nxd7 removes one of the defenders of the knight on f6. After the expected Nh5, \
White can capture the bishop on g6."

BAD (later atoms lose context):
- "Nxd7 removes one of the defenders of the knight on f6."
- "After Nh5, White can capture the bishop on g6."  <- Nh5 from where?

GOOD (each carries full move sequence from root):
- "Nxd7 removes one of the defenders of the knight on f6."
- "After Nxd7 Nh5, White can capture the bishop on g6."

CRITICAL: For consequences of move sequences, ALWAYS include the complete sequence.

PLAN ATOMS: When the commentary states a multi-move plan ending in a target \
break/idea, each atom must carry the full plan sequence — never refer to a \
future square in isolation (e.g., "the a-pawn advances from c5 to c4" is WRONG \
if c5 is itself a future state).

EXAMPLE — move played is a6, commentary:
"Black's plan is a6, Ba7, 0-0, which will lead to the pawn break with d5."

BAD (strips the plan sequence):
- "Black plans Ba7"
- "Black plans 0-0"
- "Black will play d5"  <- d5 from which position?

GOOD (each atom carries the plan):
- "After a6, Black's plan continues with Ba7 and 0-0."
- "After a6, Ba7, 0-0, Black can play the pawn break d5."

ALTERNATIVES: If the commentary mentions an ALTERNATIVE move to the played move \
(e.g., "Instead, White should play Nd5..."), put those claims under an \
`alternatives` entry keyed by the alternative move's SAN. Atoms under an \
alternative are automatically anchored to the position AFTER that alternative \
move — do NOT prefix them with "If White plays Nd5 instead". Just state the \
fact: "Nd5 attacks the bishop on e7".

RULES:
1. ONE claim per atom - split compound claims into separate atoms
2. Be specific — include piece names, squares, and concrete assertions
3. Group logically connected setup+consequence ONLY if they're inseparable
4. Each atom must be self-contained — carry the FULL move sequence from the root \
AND any premises/antecedents from surrounding sentences. A conditional ("X would \
support Y" when X is illegal) must include its premise or be dropped; a consequence \
of a prior move sequence must restate the sequence.
5. Ignore stylistic flourishes, stance/tone claims ("an attacking move", "a \
defensive idea", "weakens/strengthens the position", "improves/worsens the \
position", "creates pressure"), and author/player mental-state claims ("White \
feared X", "Black intended Y") — these are not position facts. Even negated \
versions ("does not weaken the position") are still stance claims and must be \
dropped.
6. Ignore generic chess philosophy not applied to this position, and all non-positional \
fluff (history, attribution, biography, opening popularity — see CRITICAL note above)

Also determine the author's OVERALL stance toward the played move itself \
(ignore alternatives, threats, and side comments):
- "good": author presents the move as decent, solid, strong, best, or otherwise \
acceptable. A "decent but not best" stance is still good.
- "bad": author presents the move as a mistake, blunder, inaccuracy, dubious, \
or otherwise wrong.
- "inconclusive": no clear stance, or purely descriptive with no quality judgement.

Handle negations correctly ("not a bad move" = good; "not the best" = good; \
"far from ideal" = bad).

Output valid JSON with claims grouped by logical assertions, plus the overall \
move polarity. Alternatives go in a separate `alternatives` block:
{
  "move_polarity": "good" | "bad" | "inconclusive",
  "claims": [
    {
      "claim_text": "Complete sentence(s) making this assertion",
      "atoms": ["Atomic fact 1", "Atomic fact 2"]
    }
  ],
  "alternatives": [
    {
      "move": "Nd5",
      "move_polarity": "good" | "bad" | "inconclusive",
      "relative_claim": "better" | "worse" | "inconclusive",
      "claims": [
        {
          "claim_text": "Complete sentence(s) about this alternative",
          "atoms": ["Atomic fact referencing the position AFTER this alt move"]
        }
      ]
    }
  ]
}

EXAMPLE OUTPUT:
{
  "move_polarity": "bad",
  "claims": [
    {
      "claim_text": "Bf3 is a blunder because it allows Black to play Bf5, skewering the queen on b1 and the rook on d1.",
      "atoms": [
        "Bf3 is a blunder",
        "Bf3 allows Black to play Bf5",
        "After Bf3 Bf5, the white queen is on b1",
        "After Bf3 Bf5, the white rook is on d1",
        "After Bf3 Bf5, the bishop on f5 skewers the queen on b1 and rook on d1"
      ]
    }
  ],
  "alternatives": [
    {
      "move": "Nd5",
      "move_polarity": "good",
      "relative_claim": "better",
      "claims": [
        {
          "claim_text": "Instead, White should have played Nd5, attacking the bishop on e7.",
          "atoms": ["Nd5 attacks the bishop on e7"]
        }
      ]
    }
  ]
}
\end{promptbox}

\begin{promptbox}[Classification prompt]
Classify chess claim atoms into semantic categories.

OUTPUT FORMAT: JSON array of category strings

CATEGORIES:
- quality: Move quality assessment (good, bad, blunder, strong, excellent, mistake)
- strategic: Subjective/unverifiable claims (pressure, coordination, most active, best square, initiative)
- comparison: Comparing alternatives (better, worse, instead, rather than)
- material: Material balance (wins material, sacrifice, captures, material advantage)
- tactic: Tactical motifs (pin, fork, skewer, discovered attack, trap)
- threat: Threats and attacks (threatens, targeting, attacks)
- positional: Strategic concepts (outpost, weak square, pawn structure, open file, center control)
- piece_placement: Piece locations (on square, to square, develops, moves to)
- defender: Defense claims (defends, protects, removes defender, undefended)
- move_sequence: Multi-move sequences (after X Y, variation, line, sequence)
- plan: Multi-move plans by same side (intends c3 d4, plans Nf3 Bg5)
- preparing: Setting up moves (preparing Nf6, ready to play d4, sets up castling)
- supports: Enabling moves (supports castling, enables Nd5, allows quick development)
- continuation_move: Same-side continuations (After O-O White can play Re1)
- prevents: Prevention claims (prevents Bg5, stops castling)
- opens_file: Opening files/diagonals (opens the e-file, opens long diagonal)
- forced_best: Claims a move is mandatory due to material/tactical necessity ("must play X", "forced to", "has no choice")
- general: Claims that don't fit any category above (use only as a last resort)

EXAMPLES:

INPUT: "Nf3 is a strong developing move"
OUTPUT: ["quality", "piece_placement"]

INPUT: "Bc5 develops to its most active square"
OUTPUT: ["strategic", "piece_placement"]

INPUT: "After O-O, White can continue with Re1 or d4"
OUTPUT: ["continuation_move", "piece_placement"]

INPUT: "This prepares Nf6 to challenge the center"
OUTPUT: ["preparing", "piece_placement", "positional"]

INPUT: "Bc5 supports quick kingside castling"
OUTPUT: ["supports"]

INPUT: "White intends c3, d4, and e5 to gain space"
OUTPUT: ["plan", "positional"]

INPUT: "h6 prevents Bg5"
OUTPUT: ["prevents"]

INPUT: "exd5 opens the e-file for the rook"
OUTPUT: ["opens_file", "positional"]

INPUT: "White must move the queen to avoid losing it"
OUTPUT: ["forced_best"]

IMPORTANT: An atom can have multiple categories. Include ALL that apply.
\end{promptbox}

\begin{promptbox}[Verification prompt]
You are a chess fact-checker verifying multiple atomic claims that all come from the \
SAME sentence of chess commentary. The claim text is provided for context — use it to \
resolve any ambiguity about move sequences, plans, or referenced positions in each atom.

We are flagging CHESS-FACTUAL ERRORS only. An atom DOES NOT HOLD only when it asserts
something that is demonstrably false on the board (illegal move, piece not where claimed,
square not attacked, wrong color, capture that didn't happen, etc.).

Claims that are merely weak, sub-optimal, debatable, or commentary-style judgements are
NOT factual errors. In particular:
- "Threatens X" where X is a legal move → HOLDS, even if X doesn't win material or
  wouldn't be the engine's choice.
- "Good/strong/solid move" at small engine disagreement → HOLDS (see quality guidance).
- Plans, ideas, intentions, "with the idea of…" → HOLDS if the stated follow-up is
  possible in the resulting position.
- Unverifiable or ambiguous or strategic claims → use claim_holds="false" AND confidence="n/a".
  Unverifiable is fine. DO NOT guess "false" when you cannot disprove the claim.

Only mark an atom as DOES NOT HOLD when a tool result directly contradicts it.

NEGATED CLAIMS: If an atom is phrased negatively ("X is not Y", "does not attack", \
"has not castled"), restate it positively in your head before deciding. A tool result \
confirming the negation means the atom HOLDS. Do not flip the sign.

POSITION CONTEXT:
Pre-move FEN: {pre_fen}
- Side to move: {pre_side_to_move}

MOVE PLAYED: {move_san} ({move_uci})
- {move_explanation}

Post-move FEN: {post_fen}
- Side to move: {post_side_to_move}
{alternative_context}
CLAIM (full sentence — use as context for every atom below):
{claim_text}

ATOMS TO VERIFY (verify each independently; number your verdicts in the same order):
{atom_list}

ATOM CLASSIFICATIONS:
{atom_classifications}

CRITICAL RULES:

1. MANDATORY TOOL USE: You MUST call at least one tool. Verify ALL atoms before outputting.

2. FEN SELECTION:
   - Move CREATES/DOES something ("Bg5 pins the knight"): POST-move FEN
   - Position BEFORE move ("knight was on g8"): PRE-move FEN
   - Alternative by same side: PRE-move FEN
   - Opponent's response to the move: POST-move FEN
   - When unclear: try both FENs

3. SAME-SIDE SEQUENCES (e.g., "Black plans ...g5 then ...g4"):
   - PREFER explicit form: try_variation(fen, [".", "g5", ".", "g4"]).
     "." is an opponent-move placeholder; the tool picks a random legal reply
     and samples 3 branches. Use this when the claim's move is also legal for
     the opposite side (pawn/knight/etc. moves to shared squares) — the
     legacy form can silently attribute the move to the wrong color.
   - Legacy form still works but is unsafe for same-side ambiguous moves.
   - Alternative: get_legal_moves(post_fen, "white") to check one side's options.
   - MULTI-BRANCH RESULTS: try_variation (and get_legal_moves with '.' slots)
     return up to 3 sampled branches, each with its own resulting_fen. If the
     atom appears to fail against ONE branch's FEN, you MUST re-check against
     the OTHER branches' resulting_fens before declaring DNH — a random '.'
     reply may have moved the very piece/square the atom is about. Possibility
     atoms ("can play X", "threatens Y") HOLD if $\ge 1$ branch supports them;
     failures in other branches are irrelevant. Forcedness atoms ("must",
     "cannot stop") need every sampled branch to agree.

4. PIECE COLORS:
   - {moving_color} pieces: {moving_color}, Opponent: {opponent_color}
   - PROTECTION: Only SAME-color pieces protect each other
   - ATTACKS: OPPOSITE-color pieces attack
   - Always verify color before claiming protect/attack

5. CAPTURES (only if this move was a capture):
   {capture_explanation}

6. CONTINUATION MOVES:
   - For "After O-O, White can play Re1": use get_legal_moves(post_fen, "white")
   - ALWAYS specify color parameter when checking continuation moves

7. ACTUAL MOVE QUALITY:
   - Quality claims ("O-O is a strong move"): use eval_move(PRE_fen, move) NOT post_fen

8. "SUPPORTS"/"ENABLES"/"ALLOWS" CLAIMS:
   - "Supports" means "makes possible" or "doesn't interfere with", NOT "makes it happen immediately"

9. CROSS-ATOM CONTEXT: Earlier atoms in the same sentence often establish the
   tactical basis for later atoms. APPLY prior tool findings; do not re-verify
   in isolation and do not forget them when a later atom's idiom depends on
   another atom's tactics. Example: atom 3 "after 5.d4 ..exd4 6.Rxe8+ wins Black's
   rook" HOLDS → atom 4 "d4 prevents Black from opening the e-file" HOLDS,
   because the only natural e-file opener (exd4) is refuted by atom 3.

STRICTNESS CALIBRATION:

Chess commentary is written for human readers, not engines. Read each claim
the way a reasonable annotator would have meant it, then decide.

STRICT — fail the atom on mismatch. These are concrete factual claims:
- Exact move notation and outcome.
- Exact identity: piece type, color, square, file/rank, direction.
- Legality: a move claimed as legal must be legal for the stated side.

CHARITABLE — accept if a reasonable annotator could have written the claim
given the line they had in mind. These are informal summaries:
- Net-outcome phrasing ("wins a piece", "loses the queen", "wins material")
  refers to the material balance after the forced or near-forced continuation
  implied by the variation, NOT the state at an arbitrary intermediate ply.
  Evaluate the line through its natural resolution (recaptures, forced
  replies) before judging.
- Generic categories ("a piece" ≈ minor piece or more; "material" ≈ any net
  gain) unless the surrounding text demands precision.
- Plan / idea / intention claims where the exact move order is illustrative
  rather than prescriptive.
- Commentary verbs are idiomatic, not literal. "impossible"/"cannot" = refuted
  (losing), NOT illegal; "forced"/"must" = only non-losing move, NOT only legal;
  "prevents" = refutes tactically, NOT makes illegal; "traps" = net material
  loss even if escape squares exist; "aimed at the king/kingside" = directed
  at that zone, NOT literal x-ray of the king square; "wins it" = net material
  gain across the forced continuation, even if intermediate captures recover
  some. Route these through FORCED-SEQUENCE TEST / eval_move, not get_legal_moves.

FORCED-SEQUENCE TEST:
A sequence is "forced" iff every alternative for the losing side leads to an
outcome AT LEAST AS BAD as the named line (by wp_loss or material).
Alternatives that are WORSE than the named line do NOT break forcedness. Do
not declare "not forced" just because other legal moves exist — check their
evaluations first (get_top_moves / eval_move).

When strict and charitable conflict, prefer the strict rule only if the
claim's wording is itself precise. Informal wording gets the charitable reading.

BEFORE MARKING "DOES NOT HOLD": restate the atom in one sentence of plain
chess English (what the annotator plainly meant), then cite the tool result
refuting THAT meaning. If only a literal/over-strict reading fails — the move
is legal but losing, the specific mate fails but the position is still lost,
the piece is pinned but ultimately won, the bishop x-rays the kingside but
not the king square — the atom HOLDS.

{type_specific_guidance}

When claim_holds is false and confidence is not "n/a", set error_category to one of:
  mistake/illegal-alternative   — an alternative move the commentary mentions is illegal in the position
  mistake/illegal-variation     — a move inside a proposed continuation line is illegal
  mistake/wrong-piece-at-square — commentary claims a piece is on a square, but the wrong piece (or nothing) is there
  mistake/counterfactual-compare — false before/after comparison: X was equally true/false before the move
  mistake/wrong-evaluation      — move quality label contradicted by engine evidence
  mistake/wrong-threat          — piece exists but the attack or defense claim is false
  mistake/wrong-file-open       — file or diagonal claimed open is actually blocked
  mistake/wrong-forced          — claimed forced response has legal alternatives
  mistake/no-tools              — verdict with no evidence or tool use
  judge_error/false-positive    — your reasoning actually confirms the atom holds; use this if claim_holds=false but evidence supports it
  mistake/other                 — none of the above
Disambiguation: "alternative" (standalone "could play X") vs "variation" (move inside a multi-move line).
Set error_category to null when claim_holds=true or confidence="n/a".

Set sampled_opponent_moves=true if reasoning mentions opponent moves filled by sampling
(dot '.' placeholders, "sampled branches", "random legal moves"), else false.

OUTPUT FORMAT — one entry per atom, in the SAME ORDER as the numbered list above:
{{
  "results": [
    {{
      "claim_holds": true / false / "error",
      "confidence": "high" / "medium" / "low" / "n/a" / "error",
      "reasoning": "Cite specific tool results. End with: 'Therefore the atom HOLDS/DOES NOT HOLD because ...'",
      "error_category": "<one of the 11 names above, or null>",
      "sampled_opponent_moves": true / false
    }},
    ...
  ]
}}

IMPORTANT: If any tool returns an error for a specific atom, use claim_holds="error" for that atom.
Your reasoning MUST cite specific tool results.
\end{promptbox}

\begin{promptbox}[Gold atom matching prompt]
MATCH_SYSTEM = """\
Compare two sets of atomic chess claims.
 
Given CANDIDATE atoms (from a generated explanation) and GOLD atoms \
(from an expert annotation), determine which gold atoms are COVERED \
by the candidate atoms.
 
A gold atom is "covered" if any candidate atom expresses the same \
factual insight, even if worded differently. Partial coverage counts \
if the core insight is present.
 
Output valid JSON:
{"results": [{"gold_atom": "...", "covered": true, \
"matching_candidate": "the matching candidate text"}, ...]}
 
Use null for matching_candidate when covered is false."""
\end{promptbox} 

\section{Experimental Details}
\label{app:experimental}

\subsection{Datasets} \label{sec:dataset_appendix}
We evaluate on three datasets, each targeting different aspects of commentary quality.
\textit{Textbook} (75 positions) is drawn from six games in Chernev's \textit{Logical Chess: Move by Move}~\cite{chernev2003logical}, a pedagogical text providing expert annotations for every move. We apply the filtering pipeline described in Section~\ref{s:atom_extraction} to remove bare notation, single-word evaluations, and generic filler, retaining 75 positions with substantive commentary from which we extract gold atoms. This dataset tests both factual correctness and conceptual coverage against high-quality human references.
\textit{Candidates50} (50 positions) is compiled from a Lichess study annotating 18 games by 15 players from the 2026 Candidates and Women's Candidates tournaments.\footnote{\url{https://lichess.org/study/Y1yXP80U}} We apply the same filtering and atom extraction pipeline to obtain gold atoms. This dataset provides a modern, tournament-level complement to the textbook annotations, testing whether models can match expert analysis of elite-level play.
\textit{Critical} (200 position-move pairs) targets factual correctness and move quality assessment without requiring reference commentary. We use Maia2~\cite{tang2024maia}, a human move-prediction model calibrated to Elo~2000, to identify 100 distinct positions where human players are predicted to play an inaccuracy with probability $>$10\%, following the Lichess accuracy definition\footnote{\url{https://lichess.org/page/accuracy}}. For each position, we pair both the best engine move and the predicted human mistake, yielding 200 (position, move) pairs from 100 underlying positions. Since no reference commentary exists for these positions, recall is not applicable; this dataset evaluates only factual correctness and quality assessment.

Together, the two gold sources span 24 games (6 textbook, 18 tournament); the 125 gold positions comprise 42 opening, 81 middlegame, and 2 endgame positions. Endgames are rare because their annotation tends to be move-by-move calculation rather than conceptual explanation, and is mostly removed by our filters.

We release the code under the MIT license and the annotations we created (gold atoms, move-quality labels, and engine data) under CC BY 4.0. Critical positions come from the Lichess open database (CC0). We do not redistribute the original annotation text from \textit{Logical Chess} or the Lichess study; the release includes a script that rebuilds it from a copy the user is entitled to use.

Although modest relative to training corpora, the benchmark is designed for evaluation rather than model training: all 325 position–move pairs support claim-level factual verification, 125 additionally include expert-verified gold atoms, and we report position-clustered bootstrap intervals for all main metrics.

\begin{revblock}\revcol
\subsection{Implementation Details}
\label{app:impl}
 
To enable reproducibility of these results, Table~\ref{tab:models} lists the exact model identifier used at every stage
of commentary generation and of the ACT-Eval pipeline.
\end{revblock}
 
\begin{table*}[t]
\revcol
\centering
\small
\begin{tabular}{lll}
\toprule
\textbf{Stage} & \textbf{Model} & \textbf{Identifier} \\
\midrule
\multirow{6}{*}{Commentary generation}
  & GPT-5.4          & \texttt{openai/gpt-5.4-20260305} \\
  & Claude Opus 4.7  & \texttt{anthropic/claude-4.7-opus-20260416} \\
  & Gemini 3.1 Pro   & \texttt{google/gemini-3.1-pro-preview} \\
  & DeepSeek V4 Pro  & \texttt{deepseek/deepseek-v4-pro} \\
  & Qwen3-32B        & \texttt{Qwen/Qwen3-32B} \\
  & Qwen3-8B         & \texttt{Qwen/Qwen3-8B} \\
\midrule
\multicolumn{3}{c}{\emph{ACT-Eval pipeline}} \\
\addlinespace[2pt]
Golden atom extraction & GPT-5.4 & \texttt{openai/gpt-5.4-20260305} \\
Stage 1: atomic decomposition & GPT-5.4 & \texttt{openai/gpt-5.4-20260305} \\
Stage 1: semantic classification & GPT-5.4-mini & \texttt{openai/gpt-5.4-mini-2026-03-17} \\
Stage 2: tool-augmented verification & GPT-5.4 & \texttt{openai/gpt-5.4-20260305} \\
Stage 2: cross-judge analysis & Gemini 3.1 Pro & \texttt{google/gemini-3.1-pro-preview} \\
Stage 3: gold atom matching & GPT-5.4 & \texttt{openai/gpt-5.4-20260305} \\
\bottomrule
\end{tabular}
\caption{Exact model identifiers used at each stage of generation and evaluation.}
\label{tab:models}
\end{table*}
 
\begin{revblock}\revcol
\paragraph{Generation configuration.}
All commentary is generated at temperature $0.3$ with a completion budget of
\texttt{max\_tokens} $=2048$ and the same base system prompt
(Appendix~\ref{app:prompts}), which instructs the model to act as a chess
instructor addressing an intermediate player and to produce concise,
position-specific prose. The model receives the board state as a FEN string
together with the move played in SAN. In the tool-enabled condition the model
may additionally call the tool suite defined in Table~\ref{tab:tool-suite}
before writing its explanation, for up to 10 rounds of tool calls; a model that has produced no commentary by then is recorded as a generation failure. These tools include engine analysis, board comprehension helpers, and legality checks. In the no-tool condition the model must rely entirely on parametric knowledge.
 
\paragraph{Evaluation configuration.}
Judge calls use temperature $0$. Stockfish 14.1 serves as the
backend engine for every tool call. In generation, we use depth-22 analysis, while verification tool calls run at depth-18.
 
\paragraph{Confidence intervals.}
All intervals reported in Table~\ref{tab:main-results} are $95\%$ bootstrap intervals
over $10{,}000$ resamples, computed at the position level so that atoms drawn
from the same commentary are resampled together.
\end{revblock}

\begin{revblock}\revcol
\subsection{Qwen3 Thinking-Mode Degeneracy}
\label{app:qwen}
 
Both Qwen3 models are run in No-Think mode in the main experiments. This
choice was made because thinking mode usually failed to produce any commentary
under the completion budget used throughout the pipeline, not because it
produced less favourable scores.
 
We tested both models with thinking enabled under the same
\texttt{max\_tokens} $=2048$ budget applied to every other model. Across
$N=50$ preliminary positions, the budget was frequently exhausted entirely
inside \texttt{<think>\ldots</think>} before any final commentary was emitted:
Qwen3-8B reached no final answer on $46/50$ positions ($92\%$), and Qwen3-32B
on $31/50$ positions ($62\%$). Because ACT-Eval requires a commentary to
evaluate, positions with no final answer cannot be scored at all.
 
A representative failure is shown below. The model spends the full budget
re-deriving the contents of a single rank from the FEN and second-guessing an
off-by-one square count, and never reaches the move under discussion.
 
\begin{quote}
\small
\textbf{Model:} Qwen3-32B, thinking mode.\\
\textbf{Move:} \texttt{Rc8}. Generation length: 5{,}090 characters.\\
\textbf{Position:} \fen{r2q1rk1\fs 1b1nbpp1\fs p1p1pn1p\fs 1p6\fs 3P3B\fs 1BN1PN2\fs PP2QPPP\fs 2RR2K1 b - - 5 13}
\end{quote}
\end{revblock}
 
\begin{lstlisting}[style=rev]
<think>
Okay, let's analyze this position. The user provided the FEN and the move
Rc8. First, I need to understand the position. [...] The move played is Rc8.
 
I should start by visualizing the board. [...] I'll use get_piece_at to see
what's on c8 before the move. The original position's c8 is empty because
the FEN shows 'r2q1rk1' for the first rank, so c8 is a rook? Wait, no, the
FEN is from White's perspective. [...]
 
Let's split each rank: 8th rank: r2q1rk1 a8: rook, b8: empty, c8: empty,
d8: queen, e8: empty, f8: rook, g8: knight, h8: king? Wait, that doesn't
add up. Let me count the pieces properly. [...] r (1) + 2 (empty) + q (1)
+ 1 (empty) + r (1) + k (1) = 1+2+1+1+1+1=7. So there's a missing square.
Maybe I made a mistake. [...]
 
[generation continues re-deriving the same rank repeatedly and is cut off
at the token limit before ever discussing Rc8]
\end{lstlisting}
 
\begin{revblock}\revcol
The failure is itself an instance of the board-comprehension weakness
documented in Section~\ref{sec:error-analysis}: the model cannot stably parse
a FEN into a board state, and additional reasoning tokens are consumed by the
resulting self-correction loop rather than by analysis of the move. Reported
Qwen3 numbers should nevertheless be read as reflecting the No-Think
configuration specifically.

\section{Judge Analysis}
\subsection{Cross-Judge Robustness}
\label{app:cross-judge}

\begin{table*}[ht]
\revcol
\centering

\small
\begin{tabular}{llccc}
\toprule
\textbf{Generator} & \textbf{Judge} & \textbf{Error Rate} & \textbf{Precision} & \textbf{Recall} \\
\midrule
\multirow{2}{*}{GPT-5.4}
    & GPT-5.4        & 9.5\,{\scriptsize[7.8, 11.2]}  & 57.3\,{\scriptsize[53.5, 61.1]} & 44.0\,{\scriptsize[37.3, 50.9]} \\
    & Gemini 3.1 Pro & 12.8\,{\scriptsize[11.0, 14.7]} & 52.2\,{\scriptsize[48.3, 56.0]} & 33.3\,{\scriptsize[27.0, 39.7]} \\
\midrule
\multirow{2}{*}{Gemini 3.1 Pro}
    & GPT-5.4        & 7.5\,{\scriptsize[5.6, 9.7]}   & 74.1\,{\scriptsize[69.9, 78.1]} & 60.6\,{\scriptsize[53.2, 68.0]} \\
    & Gemini 3.1 Pro & 3.2\,{\scriptsize[2.0, 4.5]}   & 81.4\,{\scriptsize[77.5, 85.1]} & 51.9\,{\scriptsize[44.4, 59.6]} \\
\midrule
\multirow{2}{*}{DeepSeek V4 Pro}
    & GPT-5.4        & 14.3\,{\scriptsize[12.5, 16.3]} & 57.1\,{\scriptsize[53.9, 60.3]} & 56.9\,{\scriptsize[49.8, 63.7]} \\
    & Gemini 3.1 Pro & 15.3\,{\scriptsize[13.3, 17.4]} & 57.3\,{\scriptsize[53.9, 60.8]} & 46.0\,{\scriptsize[38.8, 53.3]} \\
\bottomrule
\end{tabular}
\caption{\textbf{Cross-judge agreement.} The same generator outputs scored by two different judge models (GPT-5.4 vs.\ Gemini 3.1 Pro), on the positions both judges scored, excluding failed generations ($n{=}225$, $159$ and $211$ positions for the GPT-5.4, Gemini 3.1 Pro and DeepSeek V4 Pro generators; Recall over the gold-annotated subset). Bracketed ranges are $95\%$ position-clustered bootstrap CIs}
\label{tab:cross-judge}
\end{table*}

ACT-Eval uses GPT-5.4 as its judge, and GPT-5.4 is also one of the evaluated
generators. To test whether this introduces a same-model preference, we reran
the full pipeline with Gemini 3.1 Pro as an independent judge
(Table~\ref{tab:cross-judge}).

Both judges produce the same ordering on factual error rate: Gemini 3.1 Pro is
strongest, followed by GPT-5.4, then DeepSeek V4 Pro. Under the GPT-5.4 judge,
Gemini $7.5\%$ $<$ GPT-5.4 $9.5\%$ $<$ DeepSeek $14.3\%$; under the Gemini
judge, Gemini $3.2\%$ $<$ GPT-5.4 $12.8\%$ $<$ DeepSeek $15.3\%$. GPT-5.4 does
not rank its own commentary first under either judge, and Gemini ranks first
under both, so the central model comparison is robust to the choice of judge.
The absolute rates nevertheless differ across judges: ACT-Eval scores are
judge-calibrated rather than judge-invariant, and we report all main results
under a single judge. Section~\ref{app:residual} analyses the disagreements
between the two judges.

\subsection{Residual Judge Errors}
\label{app:residual}

Tool augmentation mitigates rather than eliminates hallucination. The judge
must still select relevant evidence, choose which tools to call, and interpret
what they return, and errors in any of these persist even when the underlying
tool output is correct. The cross-judge experiment
(Section~\ref{app:cross-judge}) makes this concrete, since Gemini 3.1 Pro is
the stronger chess model in our generation experiments and therefore a useful
second opinion on the same claims.

\paragraph{The GPT-5.4 judge misses errors.}
Our human validation study (Section~\ref{sec:human}) finds an asymmetry: when the
GPT-5.4 judge flags a claim as incorrect, expert annotators agree in
$83$--$94\%$ of cases, but agreement with the claims it accepts falls to
$14$--$57\%$. The cross-judge results are consistent with this. The Gemini
judge assigns higher error rates to both GPT-5.4 and DeepSeek commentary, and
inspection of the disagreements shows cases where it identifies factual and
variation errors that GPT-5.4 had accepted. These disagreements show that absolute error rates remain judge-dependent: Gemini identifies some errors accepted by GPT-5.4, while GPT-5.4 also rejects some valid claims.

\paragraph{The GPT-5.4 judge also rejects valid claims.}
Gemini-generated commentary receives a lower error rate from Gemini than from
GPT-5.4, and inspection finds several cases where GPT-5.4 rejected valid
claims by misinterpreting tool output. These concentrate on practical chess
statements using terms such as ``cannot prevent'', ``forced'', or ``must'',
where the intended meaning is tactical rather than literal. Consider the claim
\emph{``After \dots Nc6, White cannot prevent \dots Nb4''}. GPT-5.4 observed
that \texttt{\dots Nb4} remained legal after several sampled White replies and
used this to mark the claim incorrect --- reversing the evidence, since the
continued availability of \texttt{\dots Nb4} is what makes the claim true.
Under our verification rubric, ``cannot prevent'' means that White has no
satisfactory prevention, not that every preventive move is illegal. Gemini
instead evaluated the candidate preventive move \texttt{a3}, found that it
substantially worsens White's position, and concluded that White cannot
prevent the maneuver without a serious concession.

In both directions the deterministic tools returned correct evidence; what
failed was the verification procedure the judge chose and the reading it gave
the result. Residual errors therefore arise during claim interpretation, tool
selection, board-state tracking, and evidence aggregation, and ACT-Eval
continues to inherit model-dependent judge errors.
\end{revblock}

\subsection{Comparison with vanilla LLM-as-a-judge}
\label{app:gcc}
Prior work~\cite{kim2025bridging} on chess commentary evaluation relies on a vanilla LLM-as-a-judge framework, where the judge, given raw engine outputs and human ground-truth blurb, is prompted to score the commentary on a 1-5 scale for four metrics. Among them relevance (ensuring the text only includes information relevant to the move) and completeness (ensuring all critical points on the board are covered) are the main metrics for chess-specific evaluation.

In contrast, ACT-Eval reformulates both subjective dimensions into more deterministic, verifiable metrics. We define \textit{factual precision} (captured via claim accuracy and error rate) to replace relevance. A confidently-stated hallucination could be fully relevant to the reference comment, as we can see in Figure~\ref{fig:comparison}, the generated commentary receives a near-perfect GCC-Eval score of 4.9/5, despite ACT-Eval revealing a 66.7\% error rate where most atoms contain factually incorrect statements. In our framework, commentary is decomposed into atomic
claims, each verified by a suite of tools supported by engines and board-state, and the score reflects the fraction of claims that are factually correct. This shifts the metric from a
topical filter to a factual check.

Similarly, we replace completeness with \textit{atomic recall} against expert-verified gold atoms. Instead of relying on the judge, which lacks internal expert chess understanding, to define what is critical behind a move, we measure coverage against expert annotations through atom-to-atom comparison. We argue that high-quality commentary should do more than vaguely survey the board, it must successfully retrieve the specific, expert-verified tactical and strategic insights.

\section{More Related Work}
\label{sec:related-work}

\paragraph{Chess commentary generation.} Prior work has learned move-by-move explanations from large-scale social-forum data, incorporated neural and symbolic chess engines into generation, and jointly modeled chess policy and language \citep{jhamtani2018learning,zang-etal-2019-automated,lee2022improving,feng2023chessgpt}; GCC-Eval subsequently evaluates generated commentary using an LLM judge conditioned on reference commentary and engine analysis \citep{kim2025bridging}.

\paragraph{Text generation evaluation.} Evaluating natural language generation requires multi-dimensional metrics tailored to specific domains. Traditional linguistic metrics quantify surface-level fluency and lexical overlap against human references~\cite{papineni2002bleu,lin2004rouge,zhang2019bertscore}. But these reference-based metrics correlate poorly with human judgement on tasks that depend on factual correctness, motivating a separate line of work on factuality. Early work frames the problem as entailment or question-answering against a source~\cite{kryscinski2020evaluating,durmus2020feqa,fabbri2022qafacteval}; more recent work proposed a decompose-and-verify paradigm, in which generations are split into atomic claims and checked against retrieved evidence or with general-purpose tools~\cite{min-etal-2023-factscore,chern2025factool,wei2024long}.
Chess commentary is a highly specialized domain where these prior works cannot fully cover the necessary depth.
First, verifying factual correctness in chess requires routing each claim to the appropriate computational primitive rather than retrieving textual evidence. Second, commentary quality is not exhausted by factual correctness, it must also capture the nuanced strategic understanding behind a move.

\paragraph{LLM-as-a-judge in specialized domains.} Using LLM judges to evaluate text generation quality has become standard practice. While these judges show strong agreement with humans on general tasks~\cite{liu2023g}, their performance degrades sharply in domains requiring expert knowledge~\cite{szymanski2025limitations}. In chess, GCC-Eval~\cite{kim2025bridging} is the closest prior work; it attempts to compensate for this knowledge gap by augmenting the LLM judge with reference commentary and engine analysis to compensate for the lack of chess knowledge. However, this approach remains insufficient since the reference annotations are noisy and conflate fact with stylistic flourish, while engine analysis conveys only raw numerical analysis without deeper guidance or interpretation. In comparison, ACT-Eval addresses both shortcomings, we construct expert-verified gold atoms as reference, and constrain the judge to tool calls for verifiable fact checking.

\section{Human Validation Interface} \label{app:website}

\begin{figure*}
  \centering
  \includegraphics[width=0.8\linewidth]{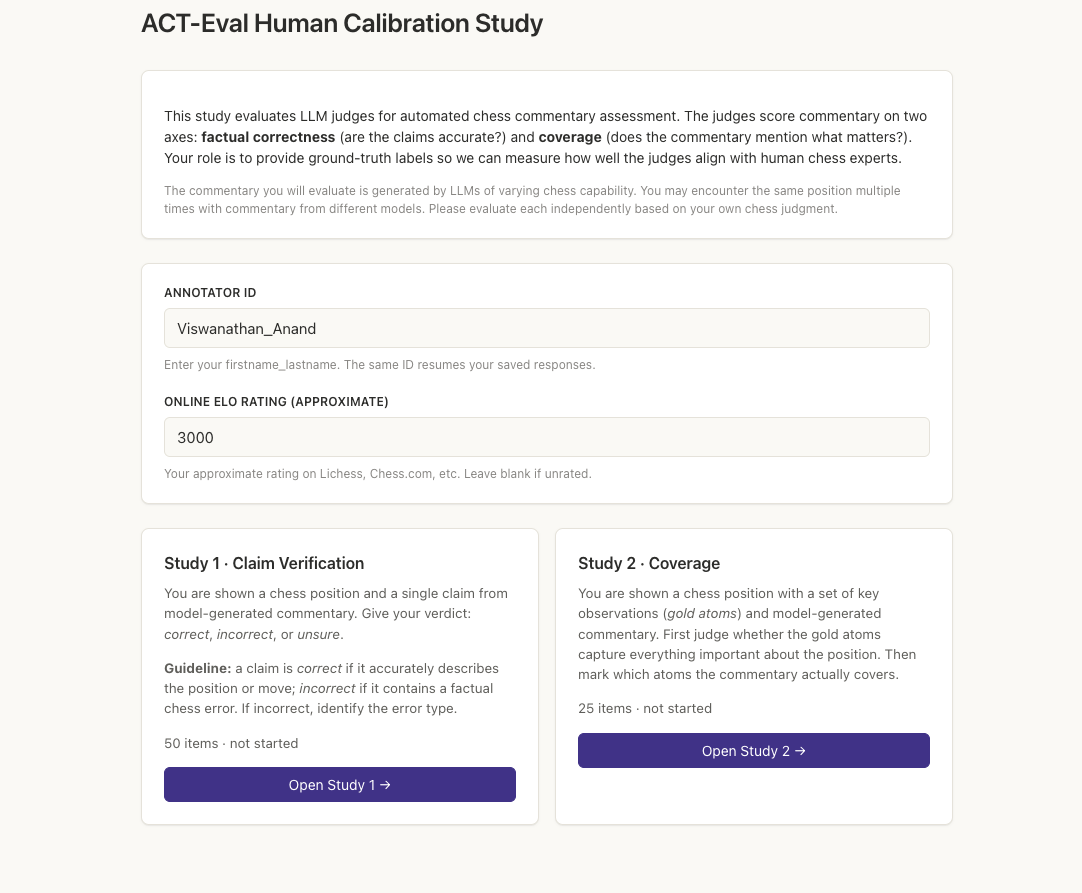}
  \caption{Screenshot of ACT-Eval Human Validation Study}
  \label{fig:website_homepage}
\end{figure*}

\begin{figure*}
  \centering
  \includegraphics[width=0.8\linewidth]{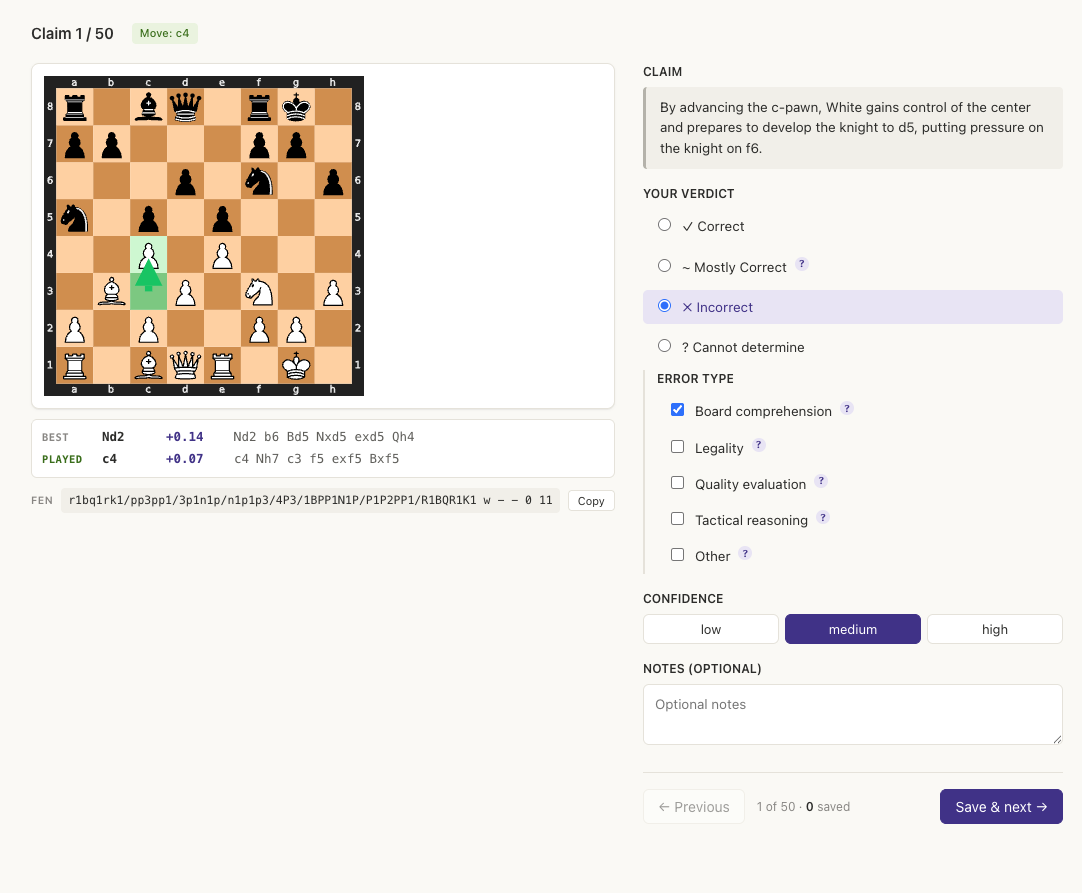}
  \caption{Screenshot of Study 1 in Human Validation}
  \label{fig:website_s1}
\end{figure*}

\begin{figure*}
  \centering
  \includegraphics[width=0.8\linewidth]{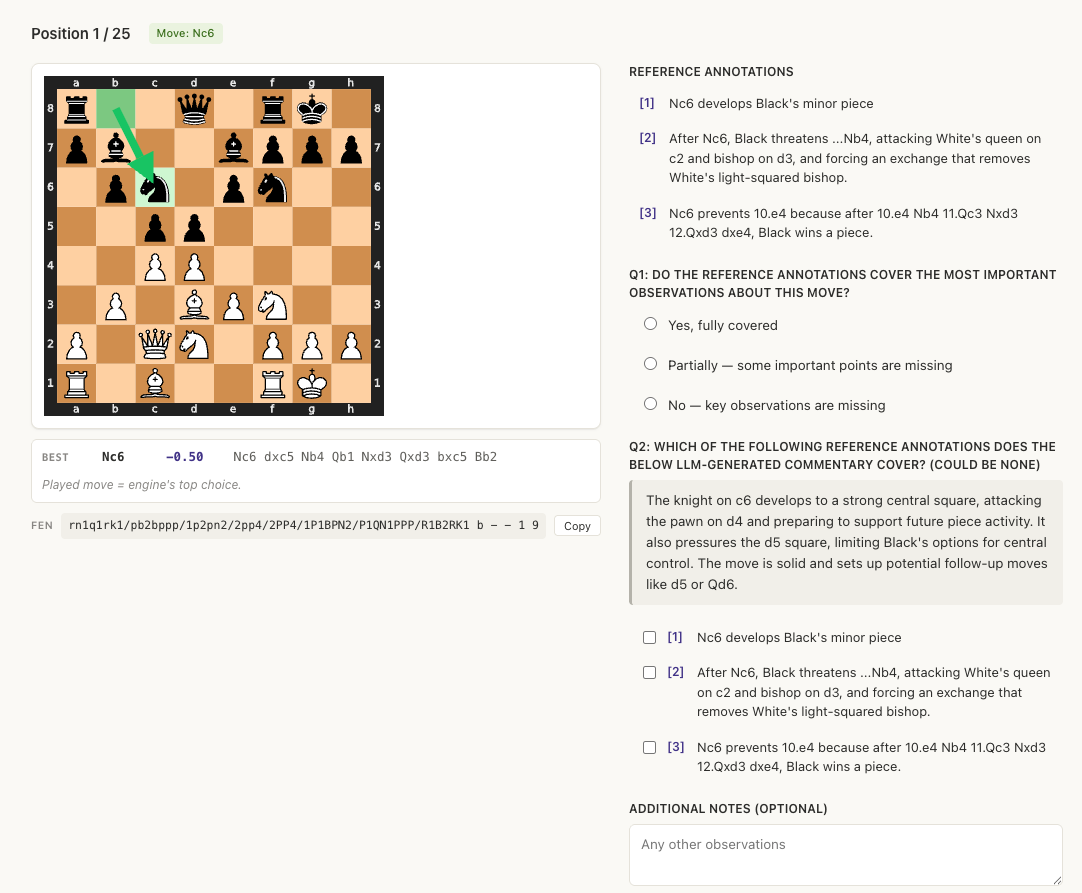}
  \caption{Screenshot of Study 2 in Human Validation}
  \label{fig:website_s2}
\end{figure*}
\label{sec:appendix}

\end{document}